\documentclass{article}

\usepackage{microtype}
\usepackage{subfigure}

\usepackage{icml2020}

\icmltitlerunning{Dark Experience for General Continual Learning}

\usepackage{times}
\usepackage{hyperref}
\usepackage{soul}
\usepackage{url}
\usepackage[utf8]{inputenc}
\usepackage[small]{caption}
\usepackage{amsmath}
\usepackage{amsthm}
\usepackage{amsbsy}
\usepackage{booktabs}
\usepackage{multicol}
\usepackage{enumitem, amssymb}
\usepackage{pifont}
\usepackage{setspace}
\usepackage[normalem]{ulem}
\usepackage{comment}
\usepackage{dsfont}
\usepackage{balance}
\usepackage{graphicx}
\usepackage{makecell}
\usepackage{cellspace}
\usepackage{marvosym}
\usepackage{bigstrut}
\usepackage{float}

\usepackage[first=0, last=9]{lcg}

\newcommand{\argmin}{\operatornamewithlimits{argmin}}
\newcommand{\argsoftmax}{\operatornamewithlimits{softmax}}
\newcommand{\KL}{\operatornamewithlimits{\textit{D}_{\textit{KL}}}}
\newcommand{\cmark}{\ding{51}}
\newcommand{\xmark}{\textbf{\textendash}}

\usepackage{multirow}
\usepackage{twoopt}
\usepackage{tikz}
\usetikzlibrary{fit}

\newcommand\tikzmark[1]{%
  \tikz[remember picture,overlay]\node[inner xsep=0pt] (#1) {};}
\newcommandtwoopt\TextboxDER[5][2.5cm][2cm]{%
\begin{tikzpicture}[remember picture,overlay]
  \coordinate (aux) at ([xshift=#1]#4);
  \node[inner ysep=5pt,yshift=0.6ex,draw=black,thick,
    fit=(#3) (aux),baseline] 
    (box) {};
\end{tikzpicture}%
}
\newcommandtwoopt\TextboxDERplusplus[5][2.5cm][2cm]{%
\begin{tikzpicture}[remember picture,overlay]
  \coordinate (aux) at ([xshift=#1]#4);
  \node[inner ysep=11pt,yshift=-0.8ex,draw=black,thick,
    fit=(#3) (aux),baseline] 
    (box) {};
\end{tikzpicture}%
}
\DeclareCaptionSubType*{algorithm}

\DeclareCaptionLabelFormat{alglabel}{Alg.~#2}

\begin{document}

\twocolumn[
\icmltitle{Dark Experience for General Continual Learning: a Strong, Simple Baseline}



\icmlsetsymbol{equal}{*}

\begin{icmlauthorlist}
    \icmlauthor{Anonymous Author}{ANINST}
\end{icmlauthorlist}

\icmlaffiliation{ANINST}{Anonymous Institution}

\icmlcorrespondingauthor{Anonymous Author}{anon.email@domain.com}

\icmlkeywords{Deep Learning, Continual Learning, Knowledge Distillation}

\vskip 0.3in
]

\printAffiliationsAndNotice{\icmlEqualContribution}

\begin{abstract}
Neural networks struggle to learn continuously, as they forget the old knowledge catastrophically whenever the data distribution changes over time. Recently, Continual Learning has inspired a plethora of approaches and evaluation settings; however, the majority of them overlooks the properties of a practical scenario, where the data stream cannot be shaped as a sequence of tasks and offline training is not viable. We work towards \textit{General Continual Learning} (GCL), where task boundaries blur and the domain and class distributions shift either gradually or suddenly. We address it through \textit{Dark Experience Replay}, namely matching the network's logits sampled throughout the optimization trajectory, thus promoting consistency with its past. By conducting an extensive analysis on top of standard benchmarks, we show that such a seemingly simple baseline outperforms consolidated approaches and leverages limited resources. To provide a better understanding, we further introduce MNIST-360, a novel GCL evaluation setting.
\end{abstract}

\section{Introduction} \label{sec:introduction}
In the last decade, the research community has mostly focused on optimizing neural networks over a finite set of shuffled data, testing on unseen examples belonging to the same distribution. Nevertheless, practical applications may face continuously changing data when new classes or tasks emerge, thus violating the i.i.d.\ property. Ideally, similarly to human beings, such models should acquire new knowledge on-the-fly and prevent the forgetting of past experiences. Still, conventional gradient-based learning paradigms lead to the catastrophic forgetting of the acquired knowledge~\cite{mccloskey1989catastrophic}. As a trivial workaround, one could store all the incoming examples and re-train from scratch when needed. However, this is often impracticable in terms of required resources. 

Continual Learning (CL) methods aim at training a neural network from a stream of non i.i.d.\ samples, relieving catastrophic forgetting whilst limiting computational costs and memory footprint~\cite{rebuffi2017icarl}. With the goal of assessing the merits of these methods, the standard experimental setting involves the creation of different tasks out of the most common image classification datasets (\textit{e.g.}~MNIST, CIFAR-10, etc.), and processing them sequentially. The authors of~\cite{hsu2018re, van2019three} identify three incremental learning (IL) settings:

\begin{figure}[t]
\centering
\includegraphics[width=0.48\textwidth]{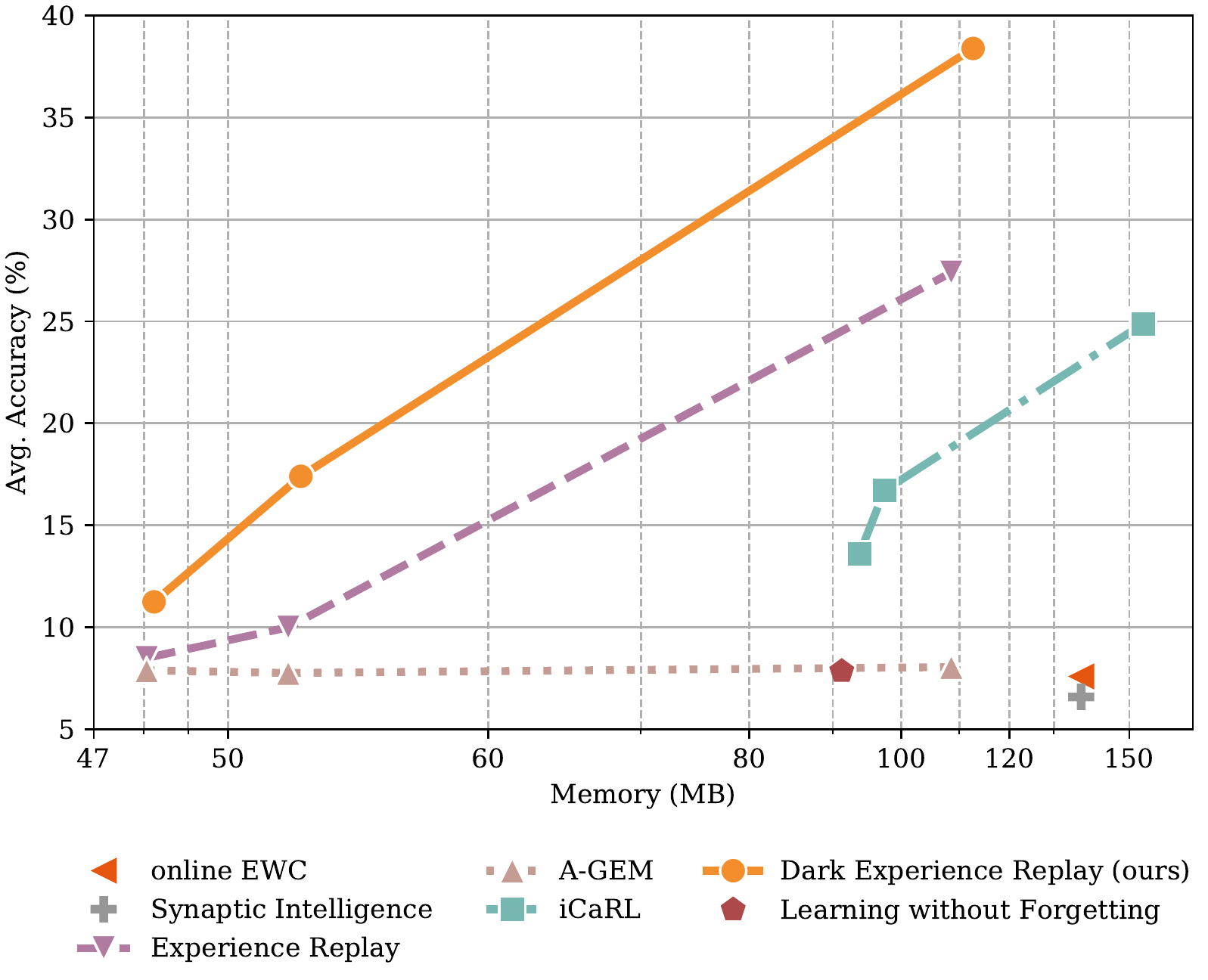}
\caption{Performance \textit{vs.}\ memory footprint for Sequential Tiny ImageNet (Class-IL). Given the approach, successive points on the same line indicate an increase in its memory buffer.}
\vspace{-1em}
\label{fig:performance_vs_memory}
\end{figure}

\begin{table}[t]
\small
\centering
\setlength\tabcolsep{2pt}
\begin{tabular}{lccccc}  
\toprule
\multirow{2}{*}{\textbf{Methods}} & \textbf{Constant} & \textbf{No task} & \textbf{Online} & \textbf{No test} \\
 & \textbf{memory} & \textbf{boundaries} & \textbf{learning} & \textbf{time oracle}\\

\midrule
PNN                  & \xmark & \xmark & \cmark & \xmark  \\
PackNet              & \xmark & \xmark & \xmark & \xmark  \\
HAT                  & \xmark & \xmark & \cmark & \xmark  \\
\midrule      
\textbf{ER}          & \cmark & \cmark & \cmark & \cmark  \\
\textbf{MER}         & \cmark & \cmark & \cmark & \cmark  \\
GEM                  & \cmark & \xmark & \cmark & \cmark  \\
A-GEM                & \cmark & \xmark & \cmark & \cmark  \\
iCaRL                & \cmark & \xmark & \xmark & \cmark  \\
\midrule      
EWC                  & \xmark & \xmark & \xmark & \cmark  \\
oEWC                 & \cmark & \xmark & \xmark & \cmark  \\
SI                   & \cmark & \xmark & \cmark & \cmark  \\
LwF                  & \cmark & \xmark & \cmark & \cmark  \\
P\&C                 & \cmark & \xmark & \xmark & \cmark  \\
\midrule
\textbf{DER (ours)}   & \cmark & \cmark & \cmark & \cmark  \\
\textbf{DER++ (ours)} & \cmark & \cmark & \cmark & \cmark  \\
\bottomrule
\end{tabular}
\caption{Continual learning approaches and their compatibility with the General Continual Learning major requirements~\cite{de2019continual}. For an exhaustive discussion, please refer to supplementary materials.}
\vspace{-0.5em}
\label{tab:gcl_requirements}
\end{table}

\begin{itemize}
\item \textbf{Task-Incremental (Task-IL)} splits the training samples into partitions of classes (tasks). At test time, task identities are provided to select the relevant classifier for each example, thus making Task-IL the most straightforward scenario.
\item \textbf{Class-Incremental (Class-IL)} divides the dataset into partitions, as Task-IL does, without providing the task identity at inference time. This is the a more challenging setting, as the network must predict both the class and the task.
\item \textbf{Domain-Incremental (Domain-IL)} feeds all classes to the network during each task, applying a task-dependent transformation on the input. Task identities remain unknown at test time.
\end{itemize}

Despite being widely adopted in literature, these protocols do not perfectly fit real-world applications: they assume to know when a distribution change occurs and tasks are unnaturally different from each other, separated by sharp boundaries. On the contrary, in many practical applications, tasks may interlace in such a way that the search for a boundary becomes infeasible. Moreover, most of the methods allow multiple iterations over the entire training set, which is not possible in an online stream with bounded memory requirements.  Recently,~\cite{de2019continual} highlighted several guidelines for General Continual Learning (GCL), in an attempt to address the issues mentioned above. GCL consists of a stream of examples, the latter shifting in their distribution and class either gradually or suddenly, providing no information about the task to be performed. Among all the CL methods, only a few can completely adhere to the requirements of such a challenging scenario (Table~\ref{tab:gcl_requirements}).

In this work, we introduce a new baseline to overcome catastrophic forgetting -- which we call \textit{Dark Experience Replay} (DER) -- suitable for the GCL setting. DER leverages Knowledge Distillation~\cite{hinton2015distilling} by matching logits of a small subset of past samples. Our proposal satisfies the major GCL guidelines: i) \textbf{online learning}: it is able to learn with no additional passages over the training data;
ii) \textbf{no task boundaries}: there is no need for a given or inferred boundary between tasks during training;
iii) \textbf{no test time oracle}: it does not require task identifiers at inference time;
iv) \textbf{constant memory}: even if the number of tasks becomes large, the memory footprint remains bounded while the model is being trained. We notice that DER complies with these requirements without affecting its performance; in fact, it outperforms most of the current state-of-the-art approaches at the cost of negligible memory overhead (as shown in Figure~\ref{fig:performance_vs_memory}). Such a claim is justified by extensive experiments encompassing the three standard CL settings.

Finally, we propose MNIST-360, a novel GCL dataset that combines the Class-IL and Domain-IL setting in the lack of explicit task boundaries. It consists of the classification of two classes of MNIST digits at a time, subject to a smooth increasing rotation. By doing so, we are further able to assess the model's ability to classify when the input distribution undergoes both swift and gradual changes.

\section{Related Work} \label{sec:related}

Continual Learning strategies have been categorized into three families~\cite{farquhar2018towards,de2019continual} of methods: \textbf{architectural methods}, \textbf{rehearsal-based methods} and \textbf{regularization-based methods}. In the following we review the most relevant works in the field and discuss whether they can be extended in the General Continual Learning setting.

\textit{\textbf{Architectural methods}} devote distinguished sets of model parameters to distinct tasks. Among these, \textbf{Progressive Neural Networks (PNN)}~\cite{rusu2016progressive} starts with a single sub-network, incrementally instantiating new ones as novel tasks occur. This methodology avoids forgetting by design, although its memory requirement grows linearly with the number of tasks.
To mitigate such an issue, \textbf{PackNet}~\cite{mallya2018packnet} and \textbf{Hard Attention to the Task (HAT)}~\cite{serra2018overcoming} share the same networks for subsequent tasks. The former prunes out unimportant weights and then re-trains the model on the current task, whereas the latter exploits task-specific attention masks to avoid interference with the past. Both of them employ a heuristic strategy to prevent intransigence, allocating additional units when needed. 

\textit{\textbf{Rehearsal-based methods}} tackle catastrophic forgetting by replaying a subset of the training data stored in a memory buffer. Early works~\cite{ratcliff1990connectionist,robins1995catastrophic} proposed \textbf{Experience Replay (ER)}, that is interleaving old samples with current data in training batches. In combination with the \textit{reservoir} strategy~\cite{vitter1985random}, this technique suits the GCL requirements, as it requires no knowledge of task boundaries. \textbf{Meta-Experience Replay (MER)}~\cite{riemer2018learning} casts replay as a meta-learning problem to maximize transfer from past tasks while minimizing interference. However, its update rule involves a high number of meta-steps for prioritizing the current example, thus making its computational cost prohibitive in an online scenario. On the other hand, \textbf{Gradient Episodic Memory (GEM)}~\cite{lopez2017gradient} and its lightweight counterpart \textbf{Averaged-GEM (A-GEM)}~\cite{chaudhry2018efficient} profit from old training data by building optimization constraints to be satisfied by the current update step. However: i) they rely on task boundaries to store examples in the memory buffer; ii) as GEM makes use of Quadratic Programming, its update step leads to similar computational limitations as MER. The authors of~\cite{rebuffi2017icarl} proposed \textbf{iCaRL}, which employs a buffer as a training set for a prototype-based \textit{nearest-mean-of-exemplars} classifier. To this end, iCaRL prevents the learnt representation from deteriorating in later tasks via a self-knowledge distillation loss term. Albeit the use of distillation is in common with our proposed baseline, we point out a clear difference: while iCaRL stores a model snapshot at each task boundary, DER plainly retains a subset of network responses for later replay.

\textit{\textbf{Regularization-based methods}} complement the loss function with additional terms to prevent forgetting. A first family of methods collects gradient information to estimate weights' importance. A loss term is then devised to prevent them from drifting too far from the previous configuration. \textbf{Elastic Weight Consolidation (EWC)}~\cite{kirkpatrick2017overcoming} accomplishes this by estimating a Fisher information matrix for each task, resulting in growing memory consumption. Conversely, \textbf{online EWC (oEWC)}~\cite{schwarz2018progress} and \textbf{Synaptic Intelligence (SI)}~\cite{zenke2017continual} rely on a single estimate. A second line of works leverages knowledge distillation for consolidating previous tasks. \textbf{Progress \& Compress (P\&C)}~\cite{schwarz2018progress} uses an \textit{Active Column} model to learn the current task. At a later stage, it distills the acquired knowledge to a \textit{Knowledge Base} model, in charge of preserving information on all previous tasks. \textbf{Learning Without Forgetting (LwF)}~\cite{li2017learning} uses an old snapshot of the model to distill its responses to current examples. As with iCaRL, this approach requires storing network weights at task boundaries, which differentiates it from the baseline we propose.

\textit{\textbf{GCL compliance.}} On top of requiring an increasing amount of memory, \textit{architectural methods} need explicit knowledge of task boundaries and task identities, thus being unsuitable for the GCL setting. These methods additionally suffer from an increasing memory footprint as more task are considered. With the exception of EWC, which shares most of the architectural methods' shortcomings, other \textit{regularization-based strategies} require a fixed memory footprint and do not need task identities. Nevertheless, they still depend on the availability of task boundaries to update their weights importance (oEWC, SI), the \textit{Knowledge Base} (P\&C), or the distillation teacher (LwF). Finally, several \textit{rehearsal-based methods} rely on task limits in order to adopt a task-stratified sampling strategy (iCaRL, GEM, A-GEM).

\section{Baseline}

In the light of the review proposed in Section~\ref{sec:related}, we draw the following remarks: on the one hand, Experience Replay provides a versatile and easy-to-implement procedure for addressing the GCL requirements at a bounded memory footprint; on the other hand, regularization-based methods deliver robust strategies to retain old knowledge. Among the latter, as observed in~\cite{li2017learning}, Knowledge Distillation proves to be especially efficient, as teacher responses encode richer information while requiring fewer resources. We combine these two methodologies in a straightforward GCL-compliant baseline, which \textit{inter alia} exhibits strong results in most standard CL settings.

Current distillation-based approaches employ task boundaries to pin a model snapshot as the new teacher, which will then provide the regularization objective during the following task. Such a practice is not suitable in the lack of sharp task limits, which is peculiar to the setting we focus on. With this in mind, we do not model the teacher in terms of its parameters at the end of a task, but rather retain its outputs (soft labels). Namely: i) we store the network responses along with their corresponding examples for replaying purposes; ii) we do it throughout the task, adopting a \textit{reservoir} strategy. Although all the methods above appoint teachers at their local optimum, our procedure does not hurt performance at all. Moreover -- even if counter-intuitive -- we experimentally observe that sampling the soft labels during the optimization trajectory delivers an interesting regularization effect (see Section~\ref{subsec:ablation}). We call the proposed baseline \textbf{Dark Experience Replay (DER)} due to its twofold nature: it relies on \textit{dark knowledge}~\cite{hinton2014dark} for distilling past memories and it remains in line with ER, leveraging its past \textit{experiences} sampled over the entire training trajectory.

\subsection{Dark Experience Replay}

Formally, a CL classification problem is split in $T$ tasks; during each task $t\in\{1,...,T\}$ input samples $x$ and their corresponding ground truth labels $y$ are drawn from an i.i.d.\ distribution $D_t$. A function $f$, with parameters $\theta$, is optimized on one task at a time in a sequential manner. We indicate the output logits with $h_\theta(x)$ and the corresponding probability distribution over the classes with $f_\theta(x) \triangleq \argsoftmax(h_\theta(x))$. At any given point in training, the goal is to learn how to correctly classify examples for the current task $t_c$, while retaining good performance on the previous ones. This corresponds to optimizing the following objective:
\begin{equation}
    \argmin_{\theta}\sum_{t=1}^{t_c}{\mathcal{L}_t},\text{where}~\mathcal{L}_t \triangleq {\mathds{E}_{(x, y) \sim D_t}\big[\ell(y, f_{\theta}(x))\big]}.
    \label{eq:obj}
\end{equation}

This is especially challenging as data from previous tasks are assumed to be unavailable, meaning that the best configuration of $\theta$ with respect to $\mathcal{L}_{t_c}$ must be sought without $D_t$ for $t\in\{1,...,{t_c}-1\}$. Ideally, we look for parameters that fit the current task well whilst approximating the behavior observed in the old ones: effectively, we encourage the network to mimic its original responses for past samples. To preserve the knowledge about previous tasks, we seek to minimize the following objective:
\begin{equation}
    \mathcal{L}_{t_c} + \alpha \sum_{t=1}^{t_c - 1}{\mathds{E}_{x \sim D_{t}}\big[\KL(f_{\theta_t^*}(x)~||~f_{\theta}(x))\big]},
    \label{eq:obj_2}
\end{equation}
where $\theta_t^*$ is the optimal set of parameters at the end of task $t$, and $\alpha$ is a hyper-parameter balancing the trade-off between the terms. This objective, which resembles the teacher-student approach, would require the availability of $D_t$ for previous tasks. To overcome such a limitation, we introduce a replay buffer $\mathcal{M}_t$ holding past \textit{experiences} for task $t$. Differently from other rehearsal-based methods~\cite{riemer2018learning}, we retain the network's logits $z \triangleq h_{\theta_t}(x)$, instead of the ground truth labels $y$.
\begin{equation}
    \mathcal{L}_{t_c} + \alpha \sum_{t=1}^{t_c - 1}{\mathds{E}_{(x,z) \sim \mathcal{M}_{t}}\big[\KL(\argsoftmax(z)~||~f_{\theta}(x))\big]} .
    \label{eq:obj_3}
\end{equation}
As we focus on General Continual Learning, we intentionally avoid relying on task boundaries to populate the buffer as the training progresses. Therefore, in place of the common task-stratified sampling strategy, we adopt \textit{reservoir} sampling~\cite{vitter1985random}: this way, we select $|\mathcal{M}|$ samples at random from the input stream, guaranteeing that they have the same probability $\frac{|\mathcal{M}|}{|\mathcal{S}|}$ of being stored in the buffer, without knowing the length of the stream $\mathcal{S}$ in advance. We can rewrite Eq.~\ref{eq:obj_3} as follows:
\begin{equation}
    \mathcal{L}_{t_c} +~\alpha~\mathds{E}_{(x,z) \sim \mathcal{M}}\big[\KL(\argsoftmax(z)~||~f_{\theta}(x))\big].
    \label{eq:obj_4}
\end{equation}
It is worth noting that this strategy implies picking logits $z$ during the optimization trajectory, so potentially far from the ones that can be observed at the task's local optimum. Although this may appear counter-intuitive, this does not hurt performance, and surprisingly it makes the optimization less prone to overfitting (see Section~\ref{subsec:ablation}). Finally, we characterize the KL divergence in Eq.~\ref{eq:obj_4} in terms of the Euclidean Distance between the corresponding logits. This adheres to~\cite{hinton2015distilling}; indeed, matching logits is a special case of distillation that avoids the information loss occurring in probability space due to the squashing function (e.g.\ softmax)~\cite{liu2018improving}. With these considerations in hands, Dark Experience Replay (DER) optimizes the following objective:
\begin{equation}
    \mathcal{L}_{t_c} +~\alpha~\mathds{E}_{(x,z) \sim \mathcal{M}}\big[ \left\lVert z - h_{\theta}(x)\right\rVert^2_2\big].
    \label{eq:obj_5}
\end{equation}
In our experiments, we approximate the expectation in Eq.~\ref{eq:obj_5} by computing gradients on a mini-batch sampled from the replay buffer.

\subsection{Dark Experience Replay++}

It is worth noting that the \textit{reservoir} strategy may weaken DER under some specific circumstances. Namely, when a distribution shift occurs in the input stream, logits that are highly biased by the training on previous tasks might be sampled for later replay. When the distribution shift is sudden, leveraging the ground truth labels could mitigate such a shortcoming. On these grounds, we additionally propose \textbf{Dark Experience Replay++ (DER++)}, which equips the objective of Eq.~\ref{eq:obj_5} with an additional term on buffer datapoints, promoting higher conditional likelihood w.r.t.\ their ground truth labels:
\begin{align}
    \mathcal{L}_{t_c} +&~\alpha~\mathds{E}_{(x',y',z') \sim \mathcal{M}}\big[ \left\lVert z' - h_{\theta}(x')\right\rVert^2_2\big]~+ \nonumber\\ +&~\beta~\mathds{E}_{(x'',y'',z'') \sim \mathcal{M}}\big[ \ell(y'', f_{\theta}(x''))\big],
\end{align}
where $\beta$ is an additional coefficient balancing the last term (DER++ collapses to DER when $\beta=0$). We provide in Algorithm~\ref{alg:combdistill} a description of DER and DER++.

\begin{algorithm}[t]
\caption{Dark Experience Replay (++)}
\label{alg:combdistill}
\begin{algorithmic}
  \STATE {\bfseries Input:} dataset $D$, parameters $\theta$, scalars $\alpha$ and $\beta$,
  \STATE \hspace{2.85em} learning rate $\lambda$
  \STATE $\mathcal{M} \gets \{\}$
  \FOR{\texttt{$D_t$} \textbf{in} $D$}
      \FOR{\texttt{$(x,y)$} \textbf{in}
          \texttt{$D_t$}}
          \STATE $(x', z', y') \gets \text{\textit{sample}}(\mathcal{M})$
          \STATE $(x'', z'', y'') \gets \text{\textit{sample}}(\mathcal{M})$
          \STATE $z \gets h_\theta(x)$ \hspace{7.5em} \textbf{DER update rule}
          \vspace{1mm}
          \STATE \tikzmark{start4} \ $\theta \gets \theta + \lambda \cdot \nabla_{\theta}[\ell(y, f_\theta(x)) +\alpha\left\lVert z' - h_\theta(x') \right\rVert^2_2~]$ \tikzmark{end4}
          \vspace{2mm}
          \STATE \hspace{11.5em} \textbf{DER++ update rule}
          \vspace{1mm}
          \STATE \tikzmark{start5} \ $\theta \gets \theta + \lambda \cdot \nabla_{\theta}[\ell(y, f_\theta(x)) +\alpha\left\lVert z' - h_\theta(x') \right\rVert^2_2~]$ \tikzmark{end5}
          \STATE \hspace{2.5em} $+~\beta~\ell(y'', f_\theta(x''))]$
          \vspace{1mm}
          \STATE $\mathcal{M} \gets \text{\textit{reservoir}}(\mathcal{M}, (x, z, y))$
      \ENDFOR
  \ENDFOR
\end{algorithmic}
\TextboxDER[0]{start4}{end4}{}
\TextboxDERplusplus[0]{start5}{end5}{}
\end{algorithm}

\section{Experiments}
\label{sec:experiments}

In this section, we draw a comprehensive comparison of our proposed baselines and several recent methods presented in Section~\ref{sec:related}. In more detail: i) we introduce a set of well known Continual Learning benchmarks in Section~\ref{subsec:cl_benchmarks}; ii) we define the evaluation protocol in Section~\ref{subsec:cl_eval_prot}; iii) we present an extensive performance analysis on the aforementioned tasks in Section~\ref{subsec:cl_exps}. Moreover, in Section~\ref{subsec:gcl_exp_setup}, we propose a novel benchmark (MNIST-360) aimed at modeling the peculiarities of General Continual Learning. By doing so, we provide an assessment of the suitable methods: to the best of our knowledge, this is the first setting catering to the GCL desiderata presented in Section~\ref{sec:introduction}. Eventually, in Section~\ref{subsec:ablation} we focus on the features of our proposal in terms of qualitative properties and resource consumption.

\subsection{Benchmarks} 
\label{subsec:cl_benchmarks}

As highlighted in Section~\ref{sec:introduction}, typical Continual Learning experiments fall into Task, Class and Domain-IL. In the following, we introduce the standard benchmarks we employ to draw a clear comparison between the proposed baselines and other well-known continual learning methods.

We conduct experiments on the Domain-Incremental scenario, leveraging two common protocols built upon the MNIST dataset~\cite{lecun1998gradient}, namely \textbf{Permuted MNIST}~\cite{kirkpatrick2017overcoming} and \textbf{Rotated MNIST}~\cite{lopez2017gradient}. They both require the learner to classify all MNIST digits for $20$ subsequent tasks. To differentiate the tasks, the former applies a random permutation to the pixels, whereas the latter rotates the images by a random angle within the interval $[0, \pi)$.

Secondly, for the Class-Incremental and Task-Incremental scenarios, we make use of three increasingly harder datasets, presenting samples belonging to different classes from MNIST, CIFAR-10~\cite{krizhevsky2009learning} and Tiny ImageNet~\cite{tinyimgnet} in a progressive fashion. In more detail, \textbf{Sequential MNIST} and \textbf{Sequential CIFAR-10} consist of $5$ tasks, each of which introduces two new classes; \textbf{Sequential Tiny ImageNet} presents $20$ distinct image classes for 10 tasks. In each setting, we show the same classes in the same fixed order across runs.

\begin{table*}[t]
\centering
{
\setlength{\tabcolsep}{2.0pt}
\small
\begin{tabular}{clcccccccc}
\toprule
 \multirow{2}{*}{\textbf{Buffer}} & \multirow{2}{*}{\textbf{Method}} & \multicolumn{2}{c}{\textbf{S-CIFAR-10}} & \multicolumn{2}{c}{\textbf{S-Tiny-ImageNet}} & \textbf{P-MNIST} & \textbf{R-MNIST}\\
\addlinespace[0.35ex]
 & & \textit{Class-IL} & \textit{Task-IL} & \textit{Class-IL} & \textit{Task-IL} & \textit{Domain-IL} & \textit{Domain-IL}\\
\midrule
\multirow{2}{*}{\xmark} & JOINT          & $92.20$\tiny{$\pm0.15$} & $98.31$\tiny{$\pm0.12$} & $59.99$\tiny{$\pm0.19$} & $82.04$\tiny{$\pm0.10$} & $94.33$\tiny{$\pm0.17$} & $95.76$\tiny{$\pm0.04$} \\
& SGD             & $19.62 $\tiny{$\pm0.05$} & $61.02 $\tiny{$\pm3.33$} & $7.92 $\tiny{$\pm0.26$} & $18.31 $\tiny{$\pm0.68$} & $40.70 $\tiny{$\pm2.33$} & $67.66 $\tiny{$\pm8.53$} \\
     \midrule

     & oEWC~\cite{schwarz2018progress}             & $19.49 $\tiny{$\pm0.12$} & $68.29 $\tiny{$\pm3.92$} & $7.58 $\tiny{$\pm0.10$} & $19.20 $\tiny{$\pm0.31$} & $\boldsymbol{75.79} $\tiny{$\boldsymbol{\pm2.25}$} & $\boldsymbol{77.35} $\tiny{$\boldsymbol{\pm5.77}$} \\
\multirow{2}{*}{\xmark} & SI~\cite{zenke2017continual}               & $19.48 $\tiny{$\pm0.17$} & $68.05 $\tiny{$\pm5.91$} & $6.58 $\tiny{$\pm0.31$} & $36.32 $\tiny{$\pm0.13$} & $65.86 $\tiny{$\pm1.57$} & $71.91 $\tiny{$\pm5.83$} \\
    & LwF~\cite{li2017learning}              & $\boldsymbol{19.61} $\tiny{$\boldsymbol{\pm0.05}$} & $63.29$\tiny{$\pm2.35$} & $\boldsymbol{8.46} $\tiny{$\boldsymbol{\pm0.22}$} & $15.85 $\tiny{$\pm0.58$} & - & - \\
     & PNN~\cite{rusu2016progressive}              & - & $\boldsymbol{95.13} $\tiny{$\boldsymbol{\pm0.72}$} & - & $\boldsymbol{67.84} $\tiny{$\boldsymbol{\pm0.29}$} & - & - \\
\midrule
    & ER~\cite{riemer2018learning} & $44.79 $\tiny{$\pm1.86$} & $91.19 $\tiny{$\pm0.94$} & $8.49 $\tiny{$\pm0.16$} & $38.17 $\tiny{$\pm2.00$} & $72.37 $\tiny{$\pm0.87$} & $85.01 $\tiny{$\pm1.90$} \\
    & GEM~\cite{lopez2017gradient} & $25.54 $\tiny{$\pm0.76$} & $90.44 $\tiny{$\pm0.94$} & - & - & $66.93 $\tiny{$\pm1.25$} & $80.80 $\tiny{$\pm1.15$} \\
    & A-GEM~\cite{chaudhry2018efficient} & $20.04 $\tiny{$\pm0.34$} & $83.88 $\tiny{$\pm1.49$} & $8.07 $\tiny{$\pm0.08$} & $22.77 $\tiny{$\pm0.03$} & $66.42 $\tiny{$\pm4.00$} & $81.91 $\tiny{$\pm0.76$} \\
    & iCaRL~\cite{rebuffi2017icarl} & $49.02$\tiny{$\pm3.20$} & $88.99 $\tiny{$\pm2.13$} & $7.53 $\tiny{$\pm0.79$} & $28.19 $\tiny{$\pm1.47$} & -     & -     \\
200 & FDR~\cite{benjamin2018measuring}         & $30.91 $\tiny{$\pm2.74$} & $91.01$\tiny{$\pm0.68$} & $8.70$\tiny{$\pm0.19$} & $40.36$\tiny{$\pm0.68$} & $74.77$\tiny{$\pm0.83$} & $85.22$\tiny{$\pm3.35$} \\
    & GSS~\cite{aljundi2019gradient}          & $39.07 $\tiny{$\pm5.59$} & $88.80 $\tiny{$\pm2.89$} & - & - & $63.72 $\tiny{$\pm0.70$} & $79.50 $\tiny{$\pm0.41$} \\
    & HAL~\cite{chaudhry2020using}          & $32.36 $\tiny{$\pm2.70$} & $82.51 $\tiny{$\pm3.20$} & - & - & $74.15 $\tiny{$\pm1.65$} & $84.02 $\tiny{$\pm0.98$} \\
    & \textbf{DER (ours)} & $61.93 $\tiny{$\pm1.79$} & $91.40 $\tiny{$\pm0.92$} & $\boldsymbol{11.87} $\tiny{$\boldsymbol{\pm0.78}$} & $40.22$\tiny{$\pm0.67$} & $81.74 $\tiny{$\pm1.07$} & $90.04 $\tiny{$\pm2.61$} \\
    & \textbf{DER++ (ours)} & $\boldsymbol{64.88} $\tiny{$\boldsymbol{\pm1.17}$} & $\boldsymbol{91.92} $\tiny{$\boldsymbol{\pm0.60}$} & $10.96 $\tiny{$\pm1.17$} & $\boldsymbol{40.87} $\tiny{$\boldsymbol{\pm1.16}$} & $\boldsymbol{83.58} $\tiny{$\boldsymbol{\pm0.59}$} & $\boldsymbol{90.43} $\tiny{$\boldsymbol{\pm1.87}$} \\
\midrule
    & ER~\cite{riemer2018learning} & $57.74 $\tiny{$\pm0.27$} & $93.61 $\tiny{$\pm0.27$} & $9.99 $\tiny{$\pm0.29$} & $48.64 $\tiny{$\pm0.46$} & $80.60 $\tiny{$\pm0.86$} & $88.91 $\tiny{$\pm1.44$} \\
    & GEM~\cite{lopez2017gradient} & $26.20 $\tiny{$\pm1.26$} & $92.16 $\tiny{$\pm0.69$} & - & - & $76.88 $\tiny{$\pm0.52$} & $81.15 $\tiny{$\pm1.98$} \\
    & A-GEM~\cite{chaudhry2018efficient} & $22.67 $\tiny{$\pm0.57$}  & $89.48 $\tiny{$\pm1.45$}  & $8.06 $\tiny{$\pm0.04$} & $25.33 $\tiny{$\pm0.49$} & $67.56 $\tiny{$\pm1.28$} & $80.31 $\tiny{$\pm6.29$} \\
    & iCaRL~\cite{rebuffi2017icarl} & $47.55$\tiny{$\pm3.95$} & $88.22$\tiny{$\pm2.62$} & $9.38 $\tiny{$\pm1.53$} & $31.55$\tiny{$\pm3.27$} & -     & -     \\
500 & FDR~\cite{benjamin2018measuring}          & $28.71$\tiny{$\pm3.23$} & $93.29$\tiny{$\pm0.59$} & $10.54$\tiny{$\pm0.21$} & $49.88$\tiny{$\pm0.71$} & $83.18$\tiny{$\pm0.53$} & $89.67 $\tiny{$\pm1.63$} \\
    & GSS~\cite{aljundi2019gradient}          & $49.73 $\tiny{$\pm4.78$} & $91.02 $\tiny{$\pm1.57$} & - & - & $76.00 $\tiny{$\pm0.87$} & $81.58 $\tiny{$\pm0.58$} \\
    & HAL~\cite{chaudhry2020using}          & $41.79 $\tiny{$\pm4.46$} & $84.54 $\tiny{$\pm2.36$} & - & - & $80.13 $\tiny{$\pm0.49$} & $85.00 $\tiny{$\pm0.96$} \\
    & \textbf{DER (ours)} & $70.51$\tiny{$\pm1.67$} & $93.40 $\tiny{$\pm0.39$} & $17.75 $\tiny{$\pm1.14$} & $51.78 $\tiny{$\pm0.88$} & $87.29 $\tiny{$\pm0.46$} & $92.24 $\tiny{$\pm1.12$} \\
    & \textbf{DER++ (ours)} & $\boldsymbol{72.70} $\tiny{$\boldsymbol{\pm1.36}$} & $\boldsymbol{93.88} $\tiny{$\boldsymbol{\pm0.50}$} &  $\boldsymbol{19.38} $\tiny{$\boldsymbol{\pm1.41}$} & $\boldsymbol{51.91} $\tiny{$\boldsymbol{\pm0.68}$} & $\boldsymbol{88.21} $\tiny{$\boldsymbol{\pm0.39}$} & $\boldsymbol{92.77} $\tiny{$\boldsymbol{\pm1.05}$} \\
\midrule
     & ER~\cite{riemer2018learning} & $82.47$\tiny{$\pm0.52$} & $\boldsymbol{96.98} $\tiny{$\boldsymbol{\pm0.17}$} & $27.40 $\tiny{$\pm0.31$} & $67.29 $\tiny{$\pm0.23$} & $89.90 $\tiny{$\pm0.13$} & $93.45 $\tiny{$\pm0.56$} \\
     & GEM~\cite{lopez2017gradient} & $25.26 $\tiny{$\pm3.46$} &  $95.55 $\tiny{$\pm0.02$} & - & - & $87.42 $\tiny{$\pm0.95$} & $88.57 $\tiny{$\pm0.40$}\\
     & A-GEM~\cite{chaudhry2018efficient} & $21.99 $\tiny{$\pm2.29$} & $90.10 $\tiny{$\pm2.09$} & $7.96 $\tiny{$\pm0.13$} & $26.22 $\tiny{$\pm0.65$} & $73.32 $\tiny{$\pm1.12$} & $80.18 $\tiny{$\pm5.52$} \\              
     & iCaRL~\cite{rebuffi2017icarl} & $55.07$\tiny{$\pm1.55$} & $92.23$\tiny{$\pm0.84$} & $14.08$\tiny{$\pm1.92$} & $40.83$\tiny{$\pm3.11$} & -     & -     \\
5120 & FDR~\cite{benjamin2018measuring}          & $19.70$\tiny{$\pm0.07$} & $94.32$\tiny{$\pm0.97$} & $28.97$\tiny{$\pm0.41$} & $68.01$\tiny{$\pm0.42$} & $90.87$\tiny{$\pm0.16$} & $94.19$\tiny{$\pm0.44$} \\
     & GSS~\cite{aljundi2019gradient}          & $67.27 $\tiny{$\pm4.27$} & $94.19 $\tiny{$\pm1.15$} & - & - & $82.22 $\tiny{$\pm1.14$} & $85.24 $\tiny{$\pm0.59$} \\
     & HAL~\cite{chaudhry2020using}          & $59.12 $\tiny{$\pm4.41$} & $88.51 $\tiny{$\pm3.32$} & - & - & $89.20 $\tiny{$\pm0.14$} & $91.17 $\tiny{$\pm0.31$} \\
     & \textbf{DER (ours)} & $83.81 $\tiny{$\pm0.33$} & $95.43 $\tiny{$\pm0.33$} & $36.73 $\tiny{$\pm0.64$} & $69.50 $\tiny{$\pm0.26$} & $91.66 $\tiny{$\pm0.11$} & $94.14 $\tiny{$\pm0.31$} \\
     & \textbf{DER++ (ours)} & $\boldsymbol{85.24} $\tiny{$\boldsymbol{\pm0.49}$} & $96.12 $\tiny{$\pm0.21$} & $\boldsymbol{39.02} $\tiny{$\boldsymbol{\pm0.97}$} & $\boldsymbol{69.84} $\tiny{$\boldsymbol{\pm0.63}$} & $\boldsymbol{92.26} $\tiny{$\boldsymbol{\pm0.17}$} & $\boldsymbol{94.65} $\tiny{$\boldsymbol{\pm0.33}$} \\
\bottomrule
\end{tabular}
}
\caption{Classification results for standard CL benchmarks, averaged across $10$ runs. `-' indicates experiments we were unable to run, because of compatibility issues (\textit{e.g.}\ between PNN, iCaRL and LwF in Domain-IL) or intractable training time (\textit{e.g.}\ GEM, HAL or GSS on Tiny ImageNet).}
\label{tab:cl_acc}
\end{table*}

\subsection{Evaluation Protocol}
\label{subsec:cl_eval_prot}

\textit{\textbf{Architecture.}}~Following ~\cite{lopez2017gradient, riemer2018learning} -- for tests we conducted on variants of the MNIST dataset -- we employ a fully-connected network with two hidden layers, each one comprising of $100$ ReLU units. Differently, as done in~\cite{rebuffi2017icarl}, we rely on ResNet18~\cite{he2016deep} (not pre-trained) for CIFAR-10 and Tiny ImageNet.

\textit{\textbf{Hyperparameter selection.}}~We select hyperparameters performing a grid-search on a validation set, the latter obtained by sampling $10\%$ of the training set. For S-MNIST, S-CIFAR-10 and S-Tiny Imagenet, we choose as best configuration the one achieving the highest average classification accuracy $\mathcal{A}_{avg}$ on all tasks both from the Class-IL and Task-IL settings, as follows:
\begin{equation}
    \mathcal{A}_{avg} \triangleq \frac{1}{2~T}\sum_{t=1}^T{\mathcal{A}_t^{\text{\textit{Task-IL}}} + \mathcal{A}_t^{\text{\textit{Class-IL}}}}.
    \label{eq:avg_acc}
\end{equation}
This does not apply to the Domain-IL setting, where we make use of the final average accuracy as the selection criterion. We always adopt the aforementioned schema to have a fair comparison among all competitors and the proposed baselines. Please refer to the supplementary materials for a detailed characterization of the hyperparameter grids we explored along with the chosen configurations.

\textit{\textbf{Training.}}~Each task lasts one epoch in all MNIST-based settings; conversely, due to their increased complexity w.r.t.\ MNIST, we set the number of epochs to $50$ for Sequential CIFAR-10 and $100$ for Sequential Tiny ImageNet. As regards batch size, we deliberately hold it out from the hyperparameter space, thus avoiding the flaw of a variable number of update steps for different methods. For the same reason, we train all the networks using the same optimization algorithm, namely Stochastic Gradient Descent (SGD).

\subsection{Experimental Results}
\label{subsec:cl_exps}

In this section, we compare our baselines\footnote{Code and experiments are fully reproducible, we make the code publicly available for research purposes.}, DER and DER++, against three regularization-based methods (\textit{oEWC}, \textit{SI}, \textit{LwF}), one architectural method (\textit{PNN}) and five rehearsal-based methods (\textit{ER}, \textit{MER}, \textit{GEM}, \textit{A-GEM}, \textit{iCaRL}). We further provide a lower bound, consisting of \textit{SGD} without any countermeasure to catastrophic forgetting. Performance have been measured in terms of average accuracy at the end of all tasks, which we in turn average across ten runs (each one involving a different random initialization). 

\textit{\textbf{Comparison with regularization and architectural methods.}}~As shown in Table~\ref{tab:cl_acc}, DER and DER++ achieve state-of-the-art performance in almost all settings, proving to be reliable in different contexts. We achieve such a robustness without compromising the intents we declared in Section~\ref{sec:introduction} concerning the GCL framework: in fact, we do not rely on the hypothesis that the data-stream resembles a sequence of tasks, which is crucial for these two families of methods. 

When compared to oEWC and SI, the gap appears unbridgeable, suggesting that regularization towards old sets of parameters does not suffice to prevent forgetting. We conjecture that this is due to a twofold reason: on the one hand, local information modeling weights importance -- computed in earlier tasks -- could become untrustworthy in later ones; on the other hand, matching logits does not preclude moving towards better configurations for past tasks. 


LwF and PNN prove to be effective in the Task-IL setting, for which they are specifically designed. However, they overtake the proposed baselines on Sequential Tiny ImageNet solely. It is worth noting that LwF suffers severe failures on Class-IL, which appears harder than the former~\cite{hsu2018re}. In this case, merely relying on a memory buffer appears a more robust strategy, as demonstrated by the results achieved by Experience Replay and its spinoffs (MER, DER, and DER++). Moreover, PNN is outright unsuited for Domain-IL and Class-IL: since it instantiates a new subnetwork for each incoming task, it is not able to handle distribution shifts within the classes nor can decide which subnetwork to use in prediction without task identifiers.

\textit{\textbf{Comparison with reharsal methods.}}~Figure~\ref{fig:cl_acc_reh} focuses on rehearsal-based methods and the benefits of a larger memory buffer\footnote{We refer the reader to the supplementary materials for a tabular representation of these results.}. As discussed above, DER and DER++ show strong performance in the majority of benchmarks, especially when dealing with small memory buffers. Interestingly, this holds especially for Domain-IL, where they outperform other approaches by a large margin at any buffer size. For these problems, a shift occurs within the input domain, but not within the classes. Hence, the relationships among them also likely persist. As an example, if it is true that during the first task number 2's visually look like 3's, this still holds when applying rotations or permutations, as it is done in the following tasks. In this regard, we argue that leveraging soft-targets in place of hard ones (as done by ER) carries more valuable information~\cite{hinton2015distilling}, exploited by DER and DER++ to preserve the similarity structure through the data-stream.

The gap in performance we observe in Domain-IL lessens when looking at Task-IL results: here, the proposed baselines perform slightly better than ER (Tiny ImageNet) or slightly worse (CIFAR-10). The underlying reason for such a difference w.r.t.\ the previous setting may be found in the way classes are observed. Indeed, as classes are presented a little at a time in exclusive subsets, the above-mentioned similarity structures arise gradually as the tasks progress; differently from Domain-IL where an overall vision is available since the first one.  

Interestingly, iCaRL achieves superior performances on Sequential TinyImagenet (200 classes) in the presence of a minimal buffer (200 samples). We believe that the reason mainly lies in the way its buffer is built. In fact, iCaRL's \textit{herding} strategy ensures all classes being equally represented in the memory buffer: for the case in the exam, each class is accordingly represented by one sample in the replay memory. On the other hand, \textit{reservoir} might omit some of them\footnote{In Tiny ImageNet, we experimentally verified that a \textit{reservoir} buffer with a size of 200 loses on average $36.7\%$ of the total number classes.}, thus potentially leading to a degradation in terms of performance. However: i) the benefits of \textit{herding} fade when increasing the buffer size; ii) it needs an additional forward pass on the last observed task, thus making it unsuitable in an online scenario; iii) most importantly, it heavily depends on task boundaries.

\subsection{MNIST-360}
\label{subsec:gcl_exp_setup}

To address the General Continual Learning desiderata, we propose a novel protocol, namely MNIST-360. It models a stream of data presenting batches of two consecutive MNIST digits (\textit{e.g.} $\{0, 1\}$, $\{1, 2\}$, $\{2, 3\}$ etc.), as depicted in Figure~\ref{fig:mnist360}. We switch the lesser of the two digits with the following one after a fixed number of steps. Aimed at modeling a continuous drift in the underlying distribution, we make sure each example of the stream to be rotated by an increasing angle w.r.t.\ the previous one. As it is impossible to distinguish sixes and nines upon rotation, we do not use nines in MNIST-360. The stream visits the nine possible couples of classes three times, allowing the model to leverage positive transfer when revisiting a previous task. In the implementation, we guarantee that: i) each example is shown once during the overall training; ii) two digits of the same class are never observed under the same rotation. We provide a detailed description of the design of training and test sets in supplementary materials.

\begin{table}[t]
    \setlength{\tabcolsep}{5pt}
    \centering
    \begin{tabular}{lccc}  
    \toprule
    \textbf{Methods} & \textbf{Buffer 200} & \textbf{Buffer 500} & \textbf{Buffer 1000}\\
    \midrule
    SGD                 & $19.09 $\tiny{$\pm0.69$} & $19.09 $\tiny{$\pm0.69$} & $19.09 $\tiny{$\pm0.69$} \\
    ER                  & $49.27 $\tiny{$\pm2.25$} & $65.04 $\tiny{$\pm1.53$} & $75.18 $\tiny{$\pm1.50$} \\
    MER                 & $48.47 $\tiny{$\pm1.60$} & $62.29 $\tiny{$\pm1.53$} & $70.01 $\tiny{$\pm1.81$} \\
    A-GEM-R              & $28.34 $\tiny{$\pm2.24$} & $28.13 $\tiny{$\pm2.62$} & $29.21 $\tiny{$\pm2.62$} \\
    \textbf{DER}        & $\boldsymbol{55.22} $\tiny{$\boldsymbol{\pm1.67}$} & $69.11 $\tiny{$\pm1.66$} & $75.97 $\tiny{$\pm2.08$} \\
    \textbf{DER++}      & $54.16 $\tiny{$\pm3.02$} & $\boldsymbol{69.62} $\tiny{$\boldsymbol{\pm1.59}$} & $\boldsymbol{76.03} $\tiny{$\boldsymbol{\pm1.61}$} \\
    \bottomrule
    \end{tabular}
    \caption{Performance of rehearsal methods on MNIST-360, in terms of classification accuracy on the test set.}
    \vspace{-1.3em}
    \label{tab:gcl_results}
\end{table}

It is worth noting that such a setting presents both sharp (change in class) and smooth (rotation) distribution shifts. Therefore, it is hard for the algorithms to identify explicit shifts in the data distribution. According to the analysis provided in Table~\ref{tab:gcl_requirements}, the suitable methods for this benchmark are ER and MER, as well as A-GEM when equipped with a reservoir memory buffer (A-GEM-R). We compare these approaches with DER and DER++, reporting the results in Table~\ref{tab:gcl_results}\footnote{We keep the same fully-connected network we used on MNIST datasets.}. As can be seen, DER and DER++ outperform the other approaches by a large margin on such a challenging scenario. Summarily, the analysis above supports the effectiveness of the proposed baselines against alternative replay methods. Due to space constraints, we refer the reader to supplementary materials for an additional evaluation regarding the memory footprint.

\begin{figure*}[t]
    \centering
    {
    \setlength\tabcolsep{2pt}
    \begin{tabular}{ccccccccc}
        \includegraphics[width=.1\textwidth]{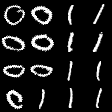} & \includegraphics[width=.1\textwidth]{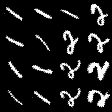} & \includegraphics[width=.1\textwidth]{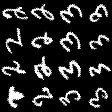} & \includegraphics[width=.1\textwidth]{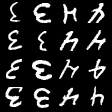} & \includegraphics[width=.1\textwidth]{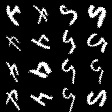} & 
        \includegraphics[width=.1\textwidth]{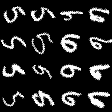} & \includegraphics[width=.1\textwidth]{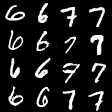} & \includegraphics[width=.1\textwidth]{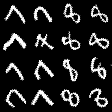} & \includegraphics[width=.1\textwidth]{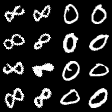} \\
        \includegraphics[width=.1\textwidth]{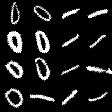} & \includegraphics[width=.1\textwidth]{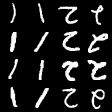} & \includegraphics[width=.1\textwidth]{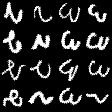} & \includegraphics[width=.1\textwidth]{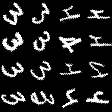} & \includegraphics[width=.1\textwidth]{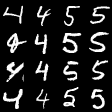} & 
        \includegraphics[width=.1\textwidth]{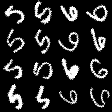} & \includegraphics[width=.1\textwidth]{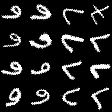} & \includegraphics[width=.1\textwidth]{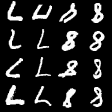} & \includegraphics[width=.1\textwidth]{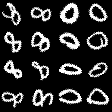} \\
    \end{tabular}
    }
    \caption{Example batches of the MNIST-360 stream.}
    \vspace{-0.5em}
    \label{fig:mnist360}
\end{figure*}

\subsection{Model analysis}
\label{subsec:ablation}

\begin{figure}[t]
    \centering
    {
    \setlength\tabcolsep{2pt}
    \begin{tabular}{cc}
        \includegraphics[width=.23\textwidth]{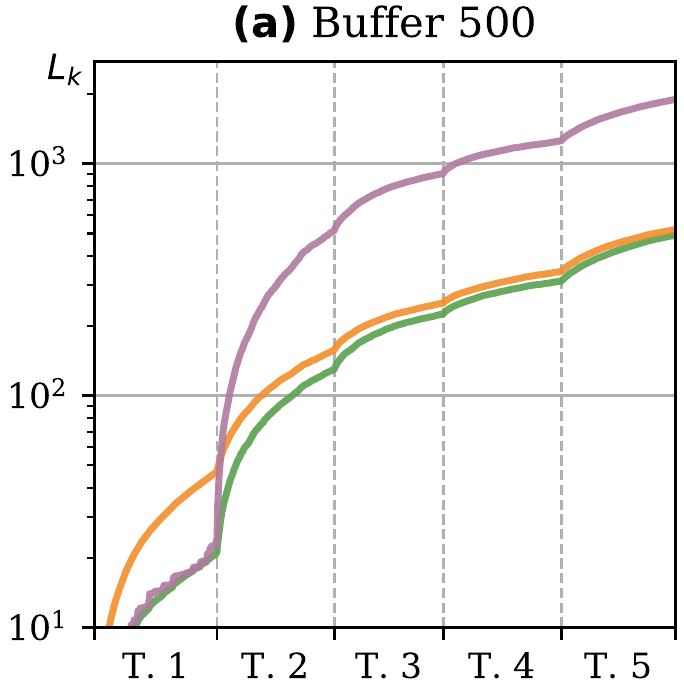} & \includegraphics[width=.23\textwidth]{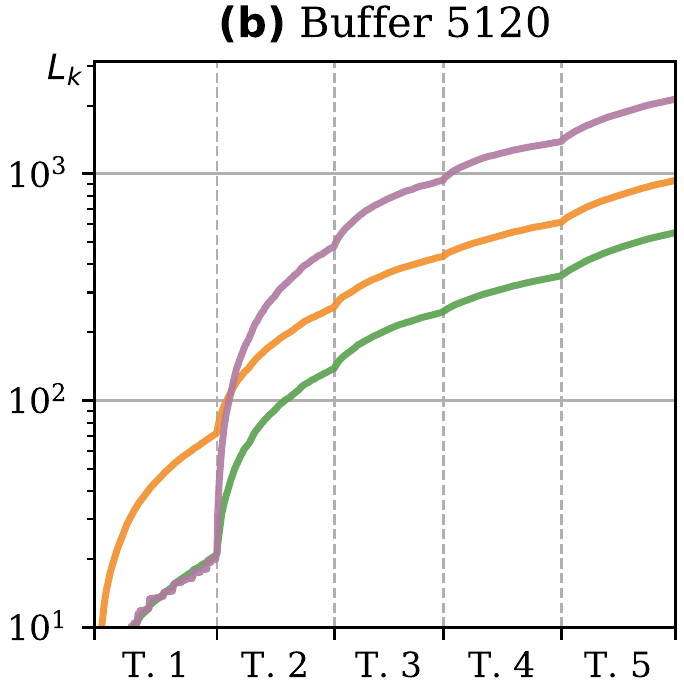} \\
        \includegraphics[width=.23\textwidth]{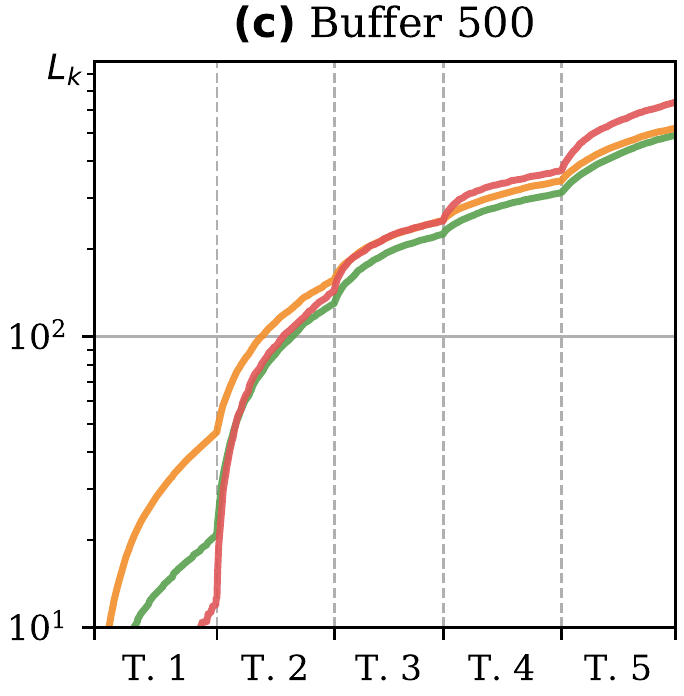} &
        \includegraphics[width=.23\textwidth]{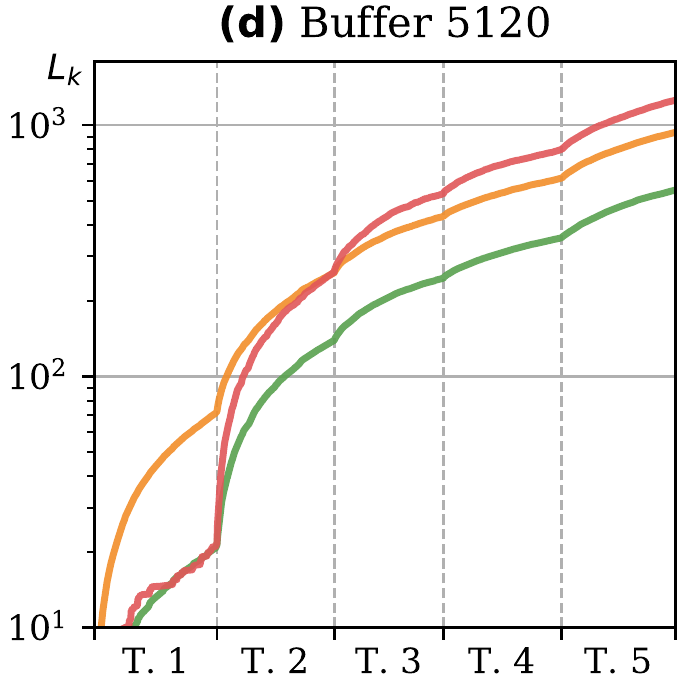}
    \end{tabular}
    \includegraphics[width=.46\textwidth]{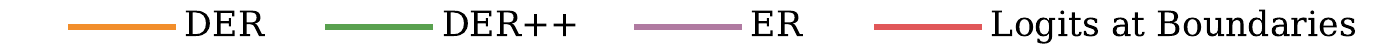} 
    }
    \caption{Trends of the weighted cumulative sum of the energy of the Fisher matrix ($L_k$), computed on S-MNIST for different buffer sizes. Lower is better. For the sake of clarity, we report it in logarithmic scale and split the analysis as follows: (a) and (b) highlight the comparison between ER and DER(++), whilst (c) and (d) focus on taking logits at the end of each task.}
    \vspace{-1.0em}
    \label{fig:abl_graph} 
\end{figure}

\textbf{\textit{On the optimization trajectory.}}~We investigate the reasons behind the regularizing capabilities of DER and DER++, conjecturing that these partially emerge when matching logits along the optimization trajectory. According to us, such an approach could impact on the strictness of soft-targets that we replay in later tasks and on its generalization properties favorably. Following this intuition, we analyze the training in terms of the weighted cumulative sum of the energy of the Fisher matrix after $k$ updates ($L_k$), as described in~\cite{liao2018approximate}. Figure~\ref{fig:abl_graph} depicts the trend of this approximation as the training on Sequential MNIST (Class-IL) progresses. As regards the comparison with ER, (a) and (b) show that its $L_k$ increases at a faster rate w.r.t.\ DER(++). According to~\cite{liao2018approximate}, this indicates that the training process leaves the optimal convergence/generalization region, thus empirically confirming the effectiveness of soft labels in continual learning scenarios. On the other hand, (c) and (d) compare sampling logits along the optimization trajectory (as done by DER(++)) against storing them at task boundaries solely. Interestingly, we observe higher generalization capabilities when exploiting reservoir.

\begin{figure}[t]
    \centering
    \includegraphics[width=0.48\textwidth]{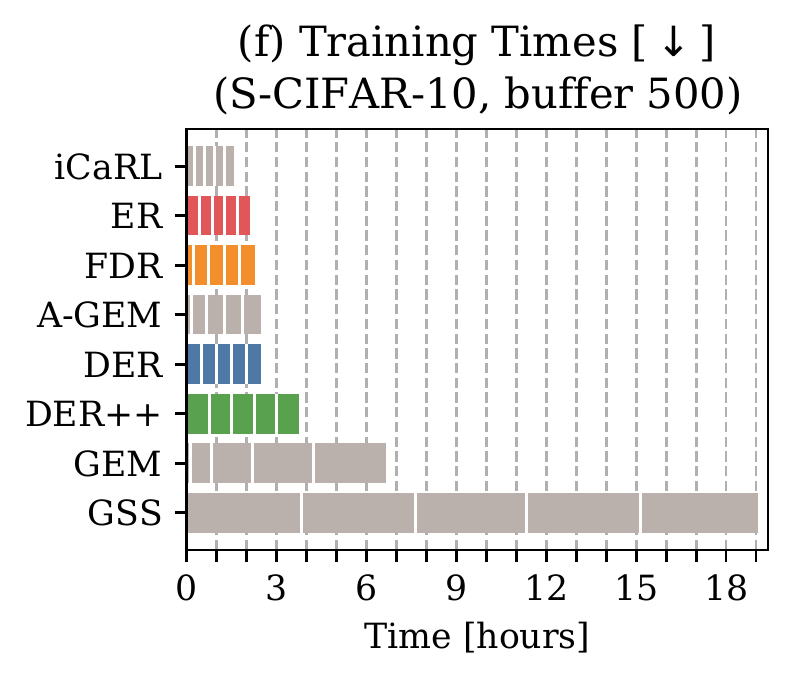}
    \caption{Wall-clock time for various rehearsal methods, the former being measured on S-MNIST.}
    \vspace{-1em}
    \label{fig:plot_seconds}
\end{figure}
\newpage

\textbf{\textit{On training time.}}~One of the goals we had in mind regards the feasibility of our proposal in a practical scenario. When facing up with a data-stream, one often cares about reducing the overall processing time: otherwise, training would not keep up with the rate at which data are made available on the stream. In this regard, we assess the performance of both DER and DER++ and other rehearsal methods in terms of wall-clock time (seconds) at the end of the last task. To guarantee a fair comparison, we conduct all tests under the same conditions, running each benchmark on a GeForce RTX 2080. Looking at Figure~\ref{fig:plot_seconds}, which reports the execution time we measured on P-MINST and S-MNIST, we draw the following remarks: i) thanks to their lightweight update rule, DER and DER++ are faster than GEM and MER; ii) the overhead between the proposed baselines and methods like ER and iCaRL appears negligible, in the face of the improvements in mitigating catastrophic forgetting.

\section{Conclusions}

In this paper, we introduce Dark Experience Replay: a simple baseline for Continual Learning, which leverages Knowledge Distillation for retaining past experience and therefore avoiding catastrophic forgetting. We demonstrate the effectiveness of our proposal through an extensive experimental analysis, carried out on top of standard benchmarks. Given the results observed in Section~\ref{subsec:ablation}, it emerges that Dark Experience provides a stronger regularization effect than plain rehearsal. At the same time, experiments highlight DER(++) as a strong baseline, which should not be ignored when assessing novel Continual Learning methods.

On the other hand, we designed it by looking ahead and adhering to the guidelines of General Continual Learning. Although being recently formalized, we argue that those provide the foundation for advances in diverse applications. Namely, the ones characterized by continuous streams of data, where relying on task switches and performing multiple passes on training data are not viable options. We therefore contributed to the study with MNIST-360 -- a novel benchmark that encompasses the peculiarities of General Continual Learning -- thus enabling future experimental studies on it.

\clearpage
\balance
\bibliographystyle{named}
\bibliography{references}

\end{document}


\section {Justification of Table~\ref{tab:gcl_requirements}}
\label{app:justifications}

Below we provide a justification for each mark of Table~\ref{tab:gcl_requirements}:
\subsection{Constant Memory}
    \begin{itemize}
        \item \textit{Distillation methods} need to accommodate a teacher model along with the current learner, at a fixed memory cost. While iCaRL maintains a snapshot of the network as teacher, LWF stores teacher responses to new task data at the beginning of each task.
        \item \textit{Rehearsal methods} need to store a memory buffer of a fixed size. This also affects iCaRL.
        \item \textit{Architectural methods} increase the model size linearly with respect to the number of tasks. In more detail, PackNet and HAT need a Boolean and float mask respectively, while PNN devotes a whole new network to each task.
        \item \textit{Regularization methods} usually require to store up to two parameters sets, thus respecting the constant memory constraint.
    \end{itemize}
\subsection{No Task Boundaries}
    \begin{itemize}
        \item \textit{Distillation methods} depend on task boundaries to appoint a new teacher. iCaRL also depends on them to update its memory buffer, in accordance with the \textit{herding} sampling strategy.
        \item \textit{Architectural methods} require to know exactly when the task finishes to update the model. PackNet also re-trains the network at this time.
        \item \textit{Regularization methods} exploit the task change to take a snapshot of the network, using it to constrain drastic changes for the most important weights (oEWC, SI). Online EWC also needs to pass over the whole training set to compute the weights importance.
        \item \textit{Rehearsal methods} can operate in the absence of task boundaries if their memory buffer exploits to the \textit{reservoir} sampling strategy. This applies to ER, GSS, MER, DER and DER++ and can easily be extended to A-GEM (by replacing \textit{ring} sampling with \textit{reservoir} as discussed in Sec.~\ref{subsec:gcl_exp_setup}). Other rehearsal approaches, however, rely on boundaries to perform specific steps: HAL hallucinates new anchors that synthesize the task it just completed, whereas FDR needs them to record converged logits to replay.
    \end{itemize}
    GEM does not strictly depend on task boundaries, but rather on task identities to associate every memorized input with its original task (as described in Sec.~$3$ of~\cite{lopez2017gradient}). This is meant to let GEM set up a separate QP for each past task (notice that this is instead unnecessary for A-GEM, which only solves one generic constraint w.r.t.\ the average gradient of all buffer items). We acknowledge that reliance on task boundaries and reliance on task identities are \textit{logically equivalent}: indeed, i) the availability of task identities straightforwardly allows any method to recognize and exploit task boundaries; ii) \textit{vice versa}, by relying on task boundaries and maintaining a task counter, one can easily associate task identities to incoming input points (under the assumption that tasks are always shown in a sequence without repetitions). This explains why Table~\ref{tab:gcl_requirements} indicates that GEM depends on task boundaries. This is also in line with what argued by the authors of~\cite{aljundi2019gradient}.
    
    \subsection{No test time oracle}
    \begin{itemize}
        \item \textit{Architectural methods} need to know the task label to modify the model accordingly before they make any prediction.
        \item LWF is designed as a multi-head method, which means that its prediction head must be chosen in accordance with the task label.
        \item \textit{Regularization methods}, \textit{rehearsal methods} and iCaRL can perform inference with no information about the task.
    \end{itemize}

\section{Details on the Implementation of MNIST-360}
\label{app:mnist360}

MNIST-360 presents the evaluated method with a sequence of MNIST digits from $0$ to $8$ shown at increasing angles.

\subsection{Training}

For Training purposes, we build batches using exemplars that belong to two consequent classes at a time, meaning that $9$ pairs of classes are possibly encountered: $(0, 1)$, $(1, 2)$, $(2, 3)$, $(3, 4)$, $(4, 5)$, $(5, 6)$, $(6, 7)$, $(7, 8)$, and $(8, 0)$. Each pair is shown in this order in $R$ rounds ($R~= 3$ in our experiments) at changing rotations. This means that MNIST-360 consists of $9 \cdot R$ pseudo-tasks, whose boundaries are not signaled to the tested method. We indicate them with $\Psi_r^{(d_1, d_2)}$ where $r~\in \{1,\ldots,R\}$ is the round number and $d_1, d_2$ are digits forming one of the pairs listed above.

As every MNIST digit $d$ appears in $2 \cdot R$ pseudo-tasks, we randomly split its example images evenly in $6$ groups $G_i^d$ where $i~\in \{1,\dots,2 \cdot R\}$. The set of exemplars that are shown in $\Psi_r^{(d_1, d_2)}$ is given as $G_{[r/2]}^{d_1} \cup G_{[r/2]+1}^{d_2}$, where $[r/2]$ is an integer division.

At the beginning of $\Psi_r^{(d_1, d_2)}$, we initialize two counters $C_{d_1}$ and $C_{d_2}$ to keep track of how many exemplars of $d_1$ and $d_2$ are shown respectively. Given batch size $B$ ($B~= 16$ in our experiments), each batch is made up of $N_{d_1}$ samples from $G_{[r/2]}^{d_1}$ and $N_{d_2}$ samples from $G_{[r/2]+1}^{d_2}$, where:
\begin{equation}
N_{d_1}~= min\left(\frac{\vert G_{[r/2]}^{d_1}\vert - C_{d_1}}{{\vert G_{[r/2]}^{d_1}\vert - C_{d_1}} + \vert G_{[r/2]+1}^{d_2}\vert - C_{d_2}}\cdot B,\vert G_{[r/2]}^{d_1}\vert - C_{d_1}\right)\\
\end{equation}
\begin{equation}
N_{d_2}~= min\left(B - N_{d_1}, \vert G_{[r/2]+1}^{d_2}\vert - C_{d_2}\right)
\end{equation}
This allows us to produce balanced batches, in which the proportion of exemplars of $d_1$ and $d_2$ is maintained the same. Pseudo-task $\Psi_r^{(d_1, d_2)}$ ends when the entirety of $G_{[r/2]}^{d_1} \cup G_{[r/2]+1}^{d_2}$ has been shown, which does not necessarily happen after a fixed number of batches.

Each digit $d$ is also associated with a counter $C^r_d$ that is never reset during training and is increased every time an exemplar of $d$ is shown to the evaluated method. Before its showing, every exemplar is rotated by 
\begin{equation}
    \frac{2\pi}{\vert d\vert}C^r_d + O_d
\label{eq:mnist360_rot}
\end{equation}
where $\vert d\vert$ is the number of total examples of digit $d$ in the training set and $O_d$ is a digit-specific angular offset, whose value for the $ith$ digit is given by $O_i~=\left(i-1\right)\frac{\pi}{2 \cdot R}$ ($O_0~=-\frac{\pi}{2 \cdot R}$, $O_1~=0$, $O_2~=\frac{\pi}{2 \cdot R}$, etc.). By so doing, every digit's exemplars are shown with an increasing rotation spanning an entire $2\pi$ angle throughout the entire procedure. Rotation changes within each pseudo-task, resulting into a gradually changing distribution. Fig.~\ref{fig:mnist360} in the main paper shows the first batch of the initial $11$ pseudo-tasks with $B=9$.

\subsection{Test}

As no task boundaries are provided, evaluation on MNIST-360 can only be carried out after the training is complete. For test purposes, digits are still shown with an increasing rotation as per Eq.~\ref{eq:mnist360_rot}, with $\vert d \vert$ referring to the test-set digit cardinality and no offset applied ($O_d=0$).

The order with which digits are shown is irrelevant, therefore no specific batching strategy is necessary and we simply show one digit at a time.

\section{Accuracy vs.\ Memory Occupation}
\begin{figure}[H]
\centering
{
\setlength\tabcolsep{5pt}
\begin{tabular}{cc}
  \includegraphics[width=.47\textwidth]{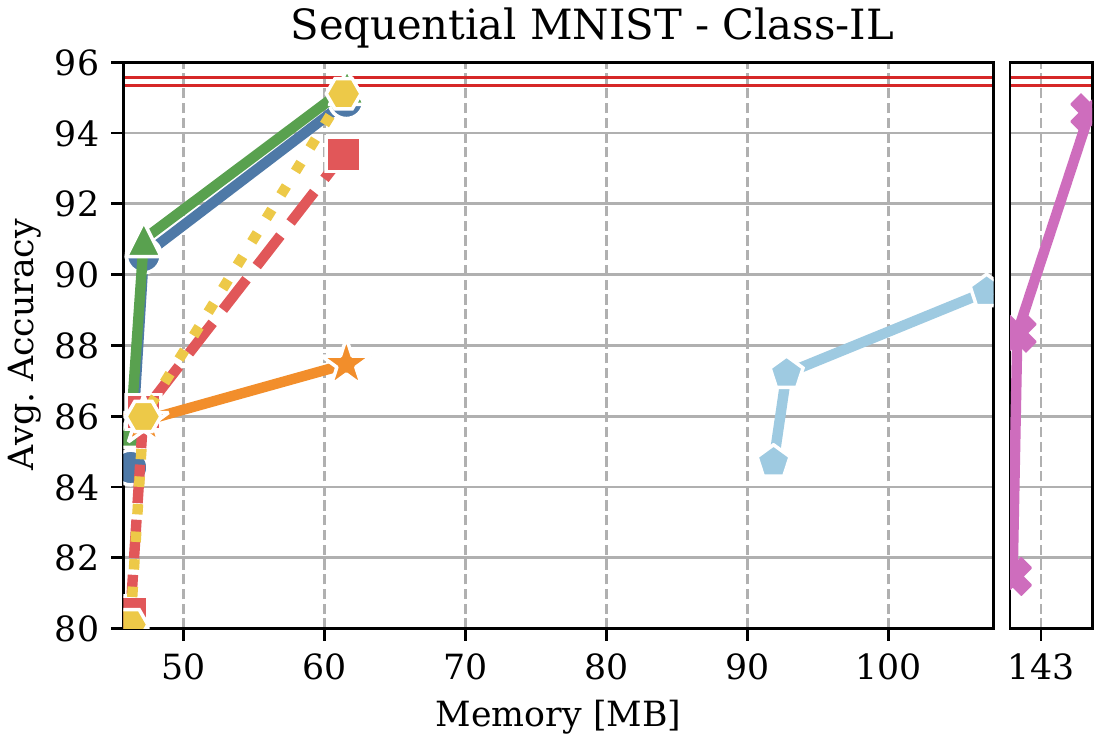} & \includegraphics[width=.47\textwidth]{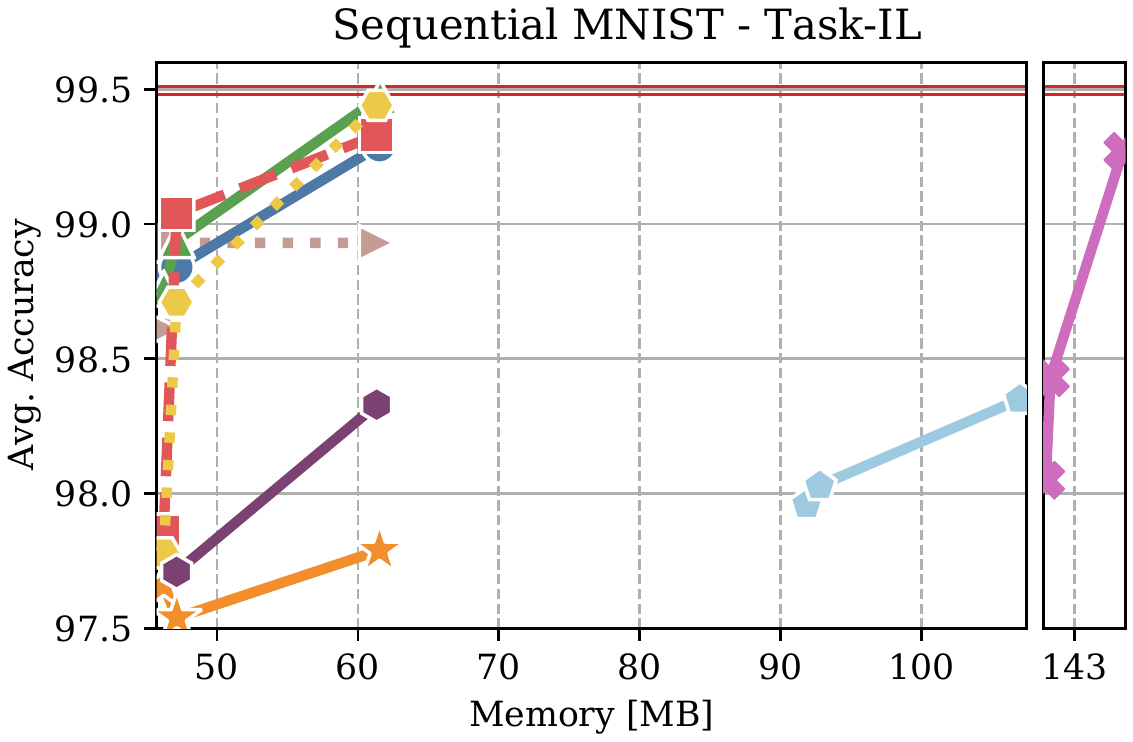} \\
\end{tabular}
}
\caption{Performance vs.\ memory allocation for the experiments of Section~\ref{sec:experiments} and~\ref{subsec:smnist}. Successive points of the same method indicate increasing buffer size. Methods with lower accuracy or excessive memory consumption may be omitted \textit{(best viewed in color).}}
\label{fig:performance_vs_memory_suppl}
\end{figure}

In Fig.~\ref{fig:performance_vs_memory_suppl}, we show how the accuracy results for the experiments in Section~\ref{subsec:cl_exps} and~\ref{subsec:smnist} relate to the total memory usage of the evaluated methods. We maintain that having a reduced memory footprint is especially important for a CL method. This is usually fairly easy to assess for rehearsal-based methods, as they clearly specify the number of items that must be saved in the memory buffer. While this could lead to the belief that they have higher memory requirements than other classes of solutions~\cite{chaudhry2018efficient}, it should be noted that architectural, distillation- and regularization-based methods can instead be characterized by non-negligible fixed overheads, making them less efficient and harder to scale.

\begin{figure}[H]
\centering
{
\setlength\tabcolsep{5pt}
\begin{tabular}{cc}
  \includegraphics[width=.47\textwidth]{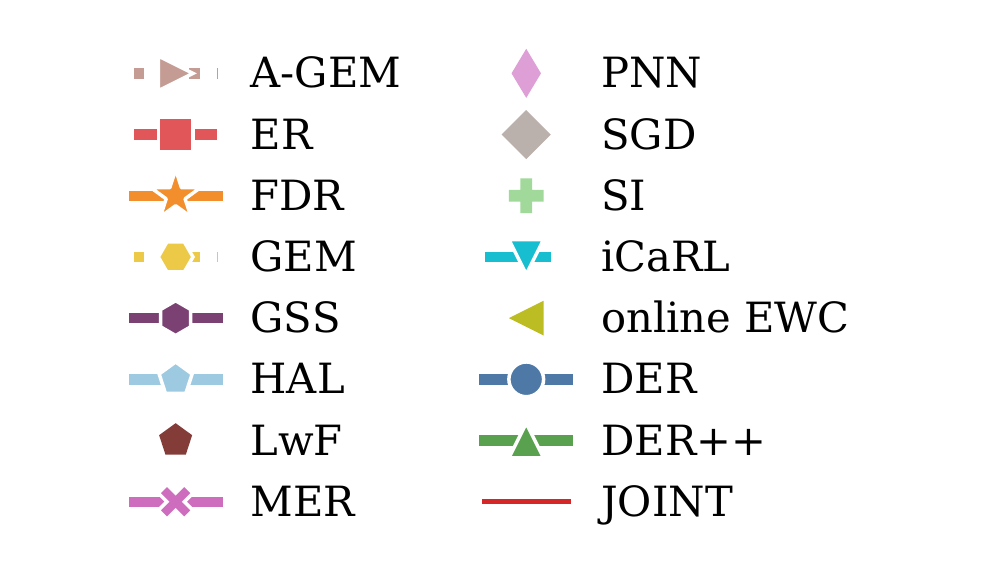} & \includegraphics[width=.47\textwidth]{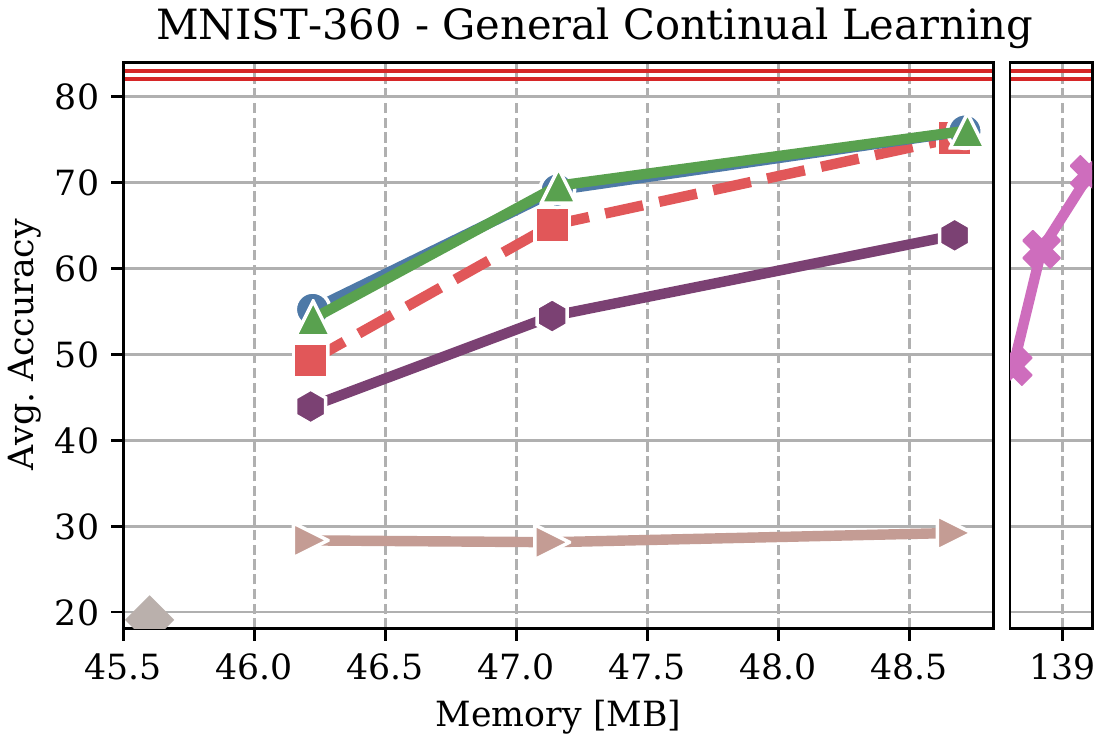} \\ 
 \end{tabular}
}
\end{figure}
\begin{figure}[H]
\centering
{
\setlength\tabcolsep{5pt}
\begin{tabular}{cc}
  \includegraphics[width=.47\textwidth]{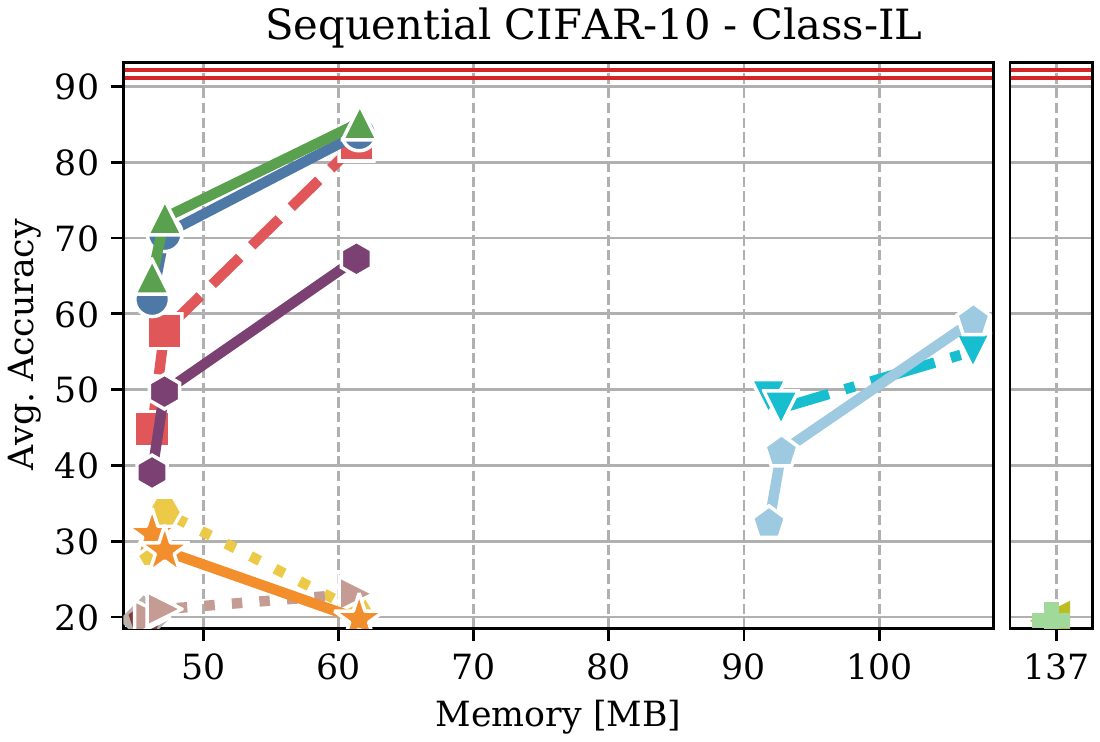} & \includegraphics[width=.47\textwidth]{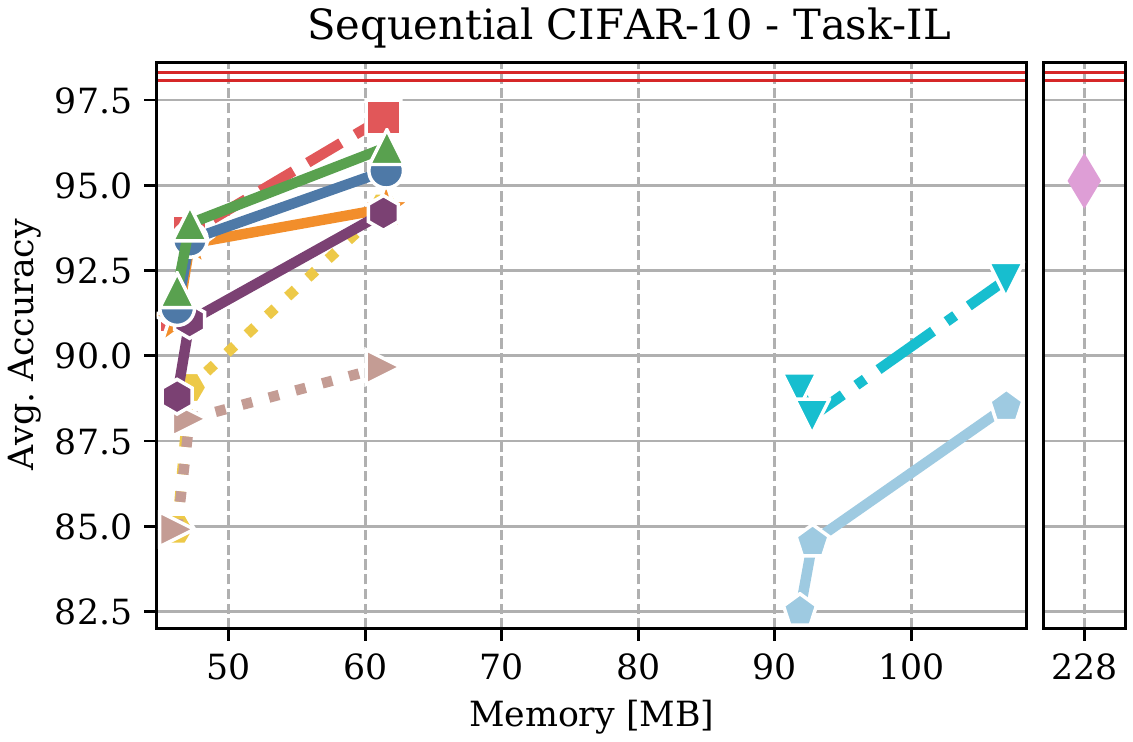} \\
 \end{tabular}
}
\end{figure}
\begin{figure}[H]
\centering
{
\setlength\tabcolsep{5pt}
\begin{tabular}{cc}
  \includegraphics[width=.47\textwidth]{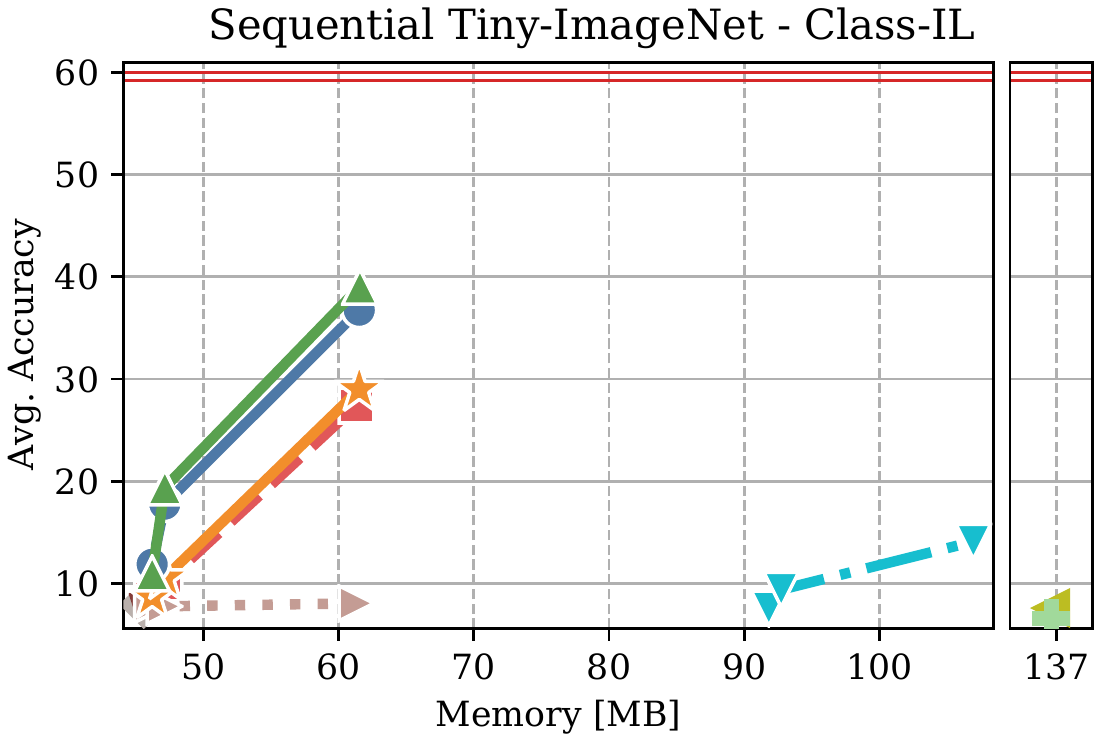} & \includegraphics[width=.47\textwidth]{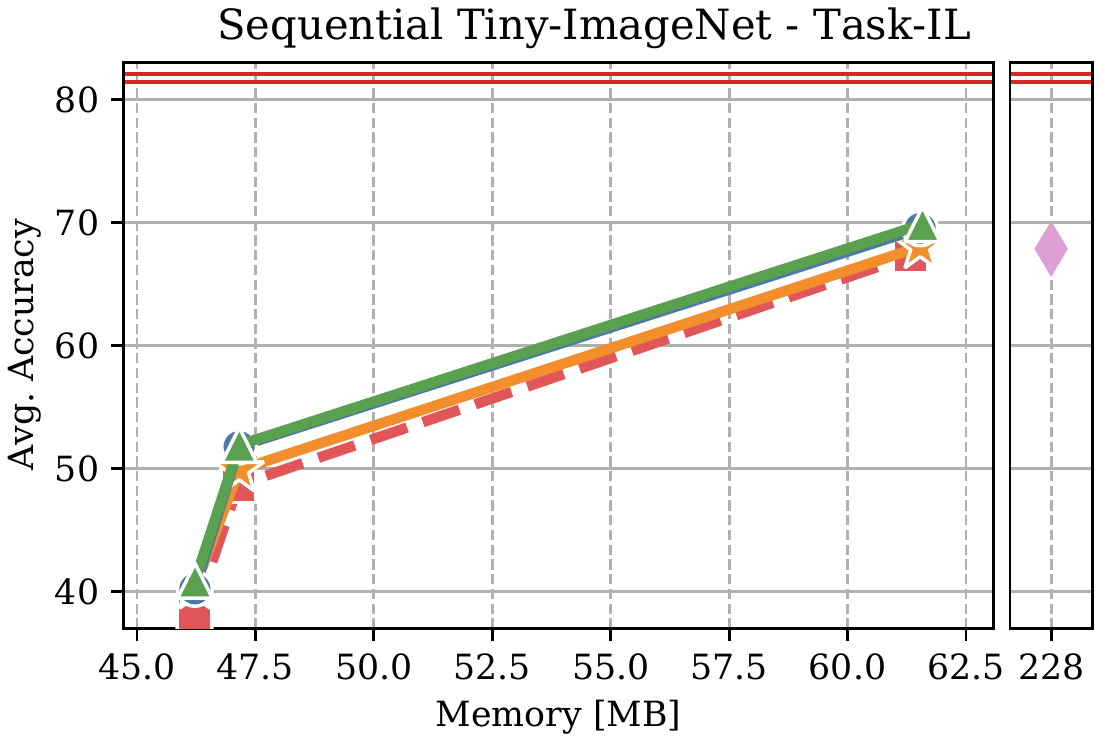} \\

\end{tabular}
}
\end{figure}
\begin{figure}[H]
\centering
{
\setlength\tabcolsep{5pt}
\begin{tabular}{cc}
  \includegraphics[width=.47\textwidth]{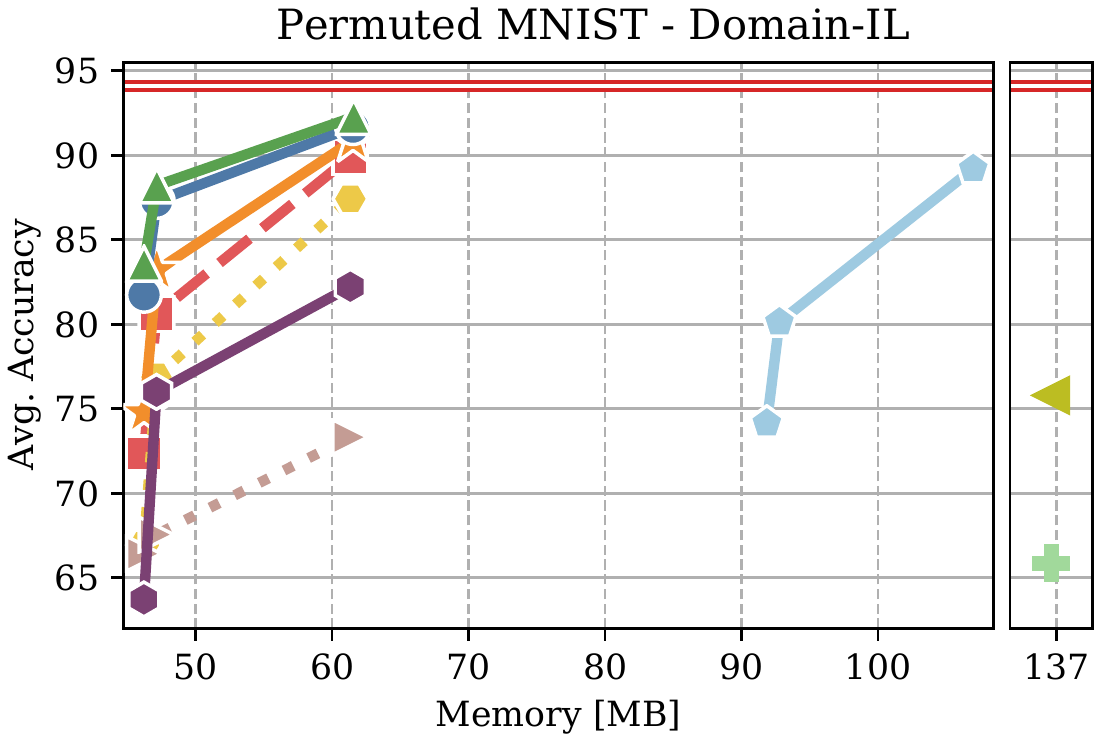} &
  \includegraphics[width=.47\textwidth]{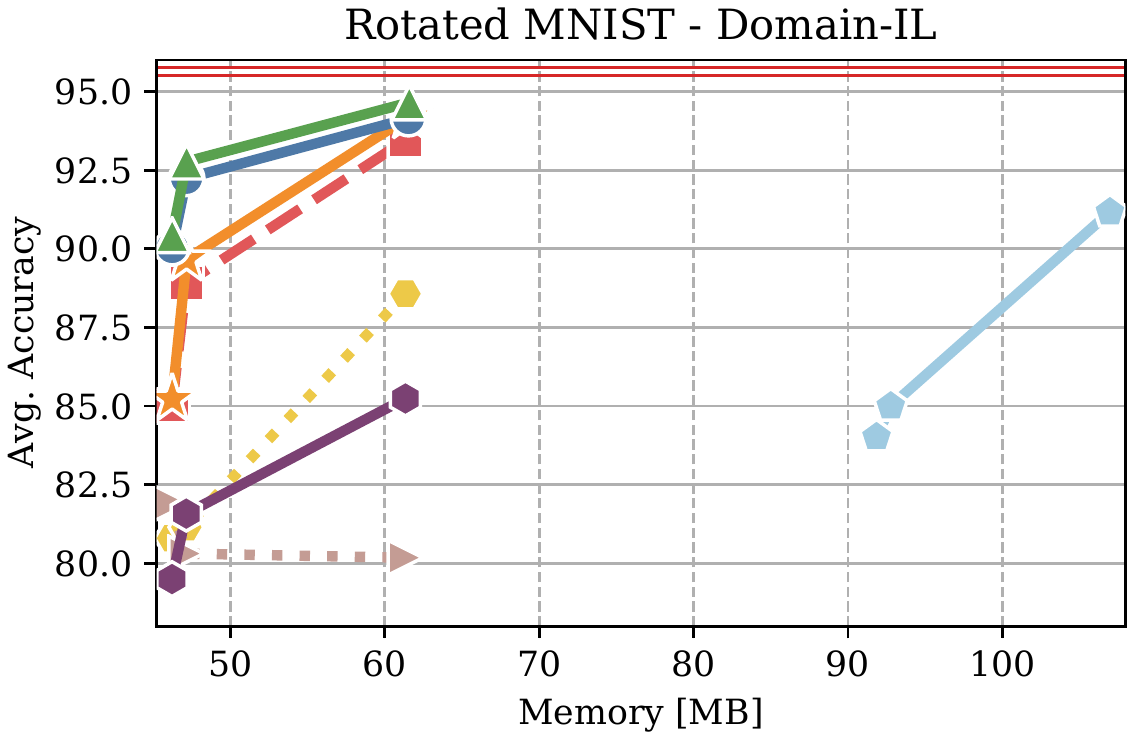} \\
 \end{tabular}
}
\end{figure}

\section{Reservoir Sampling Algorithm}
In the following, we provide the buffer insertion algorithm (\ref{alg:reservoir}) for the Reservoir Sampling strategy~\cite{vitter1985random}.
\begin{figure}[H]
    \centering
    \begin{algorithm}[H]
    \caption{- Reservoir Sampling}
    \label{alg:reservoir}
    \begin{algorithmic}
      \STATE {\bfseries Input:} memory buffer $\mathcal{M}$, number of seen examples $N$, example $x$, label $y$.
      \IF{$\mathcal{M} > N$} \STATE $\mathcal{M}[N] \gets (x,y)$ \ELSE 
      \STATE $j = \text{\textit{randomInteger}}~(\min=0,\max=N)$ 
      \IF{ $j < |\mathcal{M}|$ }
      \STATE $\mathcal{M}[j] \gets (x,y)$
      \ENDIF
      \ENDIF
      \STATE \textbf{return} $M$
    \end{algorithmic}
    \end{algorithm}
\end{figure}

\section{Details on the Implementation of iCaRL}
Although iCaRL~\cite{rebuffi2017icarl} was initially proposed for the Class-IL setting, we make it possible to use it for Task-IL as well by introducing a modification of its classification rule. Let $\mu_y$ be the average feature vector of the exemplars for class $y$ and $\phi(x)$ be the feature vector computed on example $x$, iCaRL predicts a label $y^*$ as
\begin{align}
y^\ast &= \argmin\limits_{y=1,\dots,t} \| \phi(x) - \mu_y \|.
\label{eq:icarl_classification}
\end{align}
Instead, given the tensor of average feature vectors for all classes $\boldsymbol{\mu}$, we formulate iCaRL's network response $h(x)$ as 
\begin{align}
h(x) &= - \| \phi(x) - \boldsymbol{\mu} \|.
\label{eq:icarl_new_classification}
\end{align}
Considering the argmax for $h(x)$, without masking (Class-IL setting), results in the same prediction as Eq.~\ref{eq:icarl_classification}.

It is also worth noting that iCaRL exploits a weight-decay regularization term (\textit{wd\_reg}), as suggested in~\cite{rebuffi2017icarl}, in order to make its performance competitive with the other proposed approaches.

\section {Additional Results}

\subsection{Sequential-MNIST}
\label{subsec:smnist}

Similarly to Sequential CIFAR-10, the Sequential MNIST protocol split the whole training set of the MNIST Digits dataset in $5$ tasks, each of which introduces two new digits.
\begin{table}[H]
\centering
{
\setlength{\tabcolsep}{2.0pt}
\begin{tabular}{lccccccc}
\toprule
\textbf{Model}          & \multicolumn{3}{c}{\textit{Class-IL}} &~~~~~&   \multicolumn{3}{c}{\textit{Task-IL}}  \\
\midrule
JOINT                   &                         & $95.57$\tiny{$\pm0.24$} &                         &      &                         & $99.51$\tiny{$\pm0.07$} &                       \\
SGD                     &                         & $19.60$\tiny{$\pm0.04$} &                         &      &                         & $94.94$\tiny{$\pm2.18$} &                       \\
\midrule
oEWC                    &                         & $\boldsymbol{20.46}$\tiny{$\boldsymbol{\pm1.01}$} &                         &      &                         & $98.39$\tiny{$\pm0.48$} &                       \\
SI                      &                         & $19.27$\tiny{$\pm0.30$} &                         &      &                         & $96.00$\tiny{$\pm2.04$} &                       \\
LwF                     &                         & $19.62$\tiny{$\pm0.01$} &                         &      &                         & $94.11$\tiny{$\pm3.01$} &                       \\
PNN                     &                         & -                       &                         &      &                         & $\boldsymbol{99.23}$\tiny{$\boldsymbol{\pm0.20}$} &                       \\
\midrule
\textbf{Buffer}         & 200   & 500   & 5120 &                & 200   & 500   & 5120                  \\
\midrule
ER                      & $80.43$\tiny{$\pm1.89$} & $86.12$\tiny{$\pm1.89$} & $93.40$\tiny{$\pm1.29$} &      & $97.86$\tiny{$\pm0.35$} & $\boldsymbol{99.04}$\tiny{$\boldsymbol{\pm0.18}$} & $99.33$\tiny{$\pm0.22$} \\
MER                     & $81.47$\tiny{$\pm1.56$} & $88.35$\tiny{$\pm0.41$} & $94.57$\tiny{$\pm0.18$} &      & $98.05$\tiny{$\pm0.25$} & $98.43$\tiny{$\pm0.11$} & $99.27$\tiny{$\pm0.09$} \\
GEM                     & $80.11$\tiny{$\pm1.54$} & $85.99$\tiny{$\pm1.35$} & $95.11$\tiny{$\pm0.87$} &      & $97.78$\tiny{$\pm0.25$} & $98.71$\tiny{$\pm0.20$} & $99.44$\tiny{$\pm0.12$} \\
A-GEM                   & $45.72$\tiny{$\pm4.26$} & $46.66$\tiny{$\pm5.85$} & $54.24$\tiny{$\pm6.49$} &      & $98.61$\tiny{$\pm0.24$} & $98.93$\tiny{$\pm0.21$} & $98.93$\tiny{$\pm0.20$} \\
iCaRL                   & $70.51$\tiny{$\pm0.53$} & $70.10$\tiny{$\pm1.08$} & $70.60$\tiny{$\pm1.03$} &      & $98.28$\tiny{$\pm0.09$} & $98.32$\tiny{$\pm0.07$} & $98.32$\tiny{$\pm0.11$} \\
FDR                     & $79.43$\tiny{$\pm3.26$} & $85.87$\tiny{$\pm4.04$} & $87.47$\tiny{$\pm3.15$} &      & $97.66$\tiny{$\pm0.18$} & $97.54$\tiny{$\pm1.90$} & $97.79$\tiny{$\pm1.33$} \\
GSS                     & $38.90$\tiny{$\pm2.49$} & $49.76$\tiny{$\pm4.73$} & $89.39$\tiny{$\pm0.75$} &      & $95.02$\tiny{$\pm1.85$} & $97.71$\tiny{$\pm0.53$} & $98.33$\tiny{$\pm0.17$} \\
HAL                     & $84.70$\tiny{$\pm0.87$} & $87.21$\tiny{$\pm0.49$} & $89.52$\tiny{$\pm0.96$} &      & $97.96$\tiny{$\pm0.21$} & $98.03$\tiny{$\pm0.22$} & $98.35$\tiny{$\pm0.17$} \\
\textbf{DER (ours)}     & $84.55$\tiny{$\pm1.64$} & $90.54$\tiny{$\pm1.18$} & $94.90$\tiny{$\pm0.57$} &      & $\boldsymbol{98.80}$\tiny{$\boldsymbol{\pm0.15}$} & $98.84$\tiny{$\pm0.13$} & $99.29$\tiny{$\pm0.11$} \\
\textbf{DER++ (ours)}   & $\boldsymbol{85.61}$\tiny{$\boldsymbol{\pm1.40}$} & $\boldsymbol{91.00}$\tiny{$\boldsymbol{\pm1.49}$} & $\boldsymbol{95.30}$\tiny{$\boldsymbol{\pm1.20}$} &      & $98.76$\tiny{$\pm0.28$} & $98.94$\tiny{$\pm0.27$} & $\boldsymbol{99.47}$\tiny{$\boldsymbol{\pm0.07}$} \\
\bottomrule
\end{tabular}
}
\vspace{0.5em}
\caption{Results for the Sequential-MNIST dataset.}
\end{table}

\subsection{Additional Comparisons with Experience Replay}
\label{subsec:chaudhry_mnist}

\begin{figure}[H]
\centering
\includegraphics[width=0.9\textwidth]{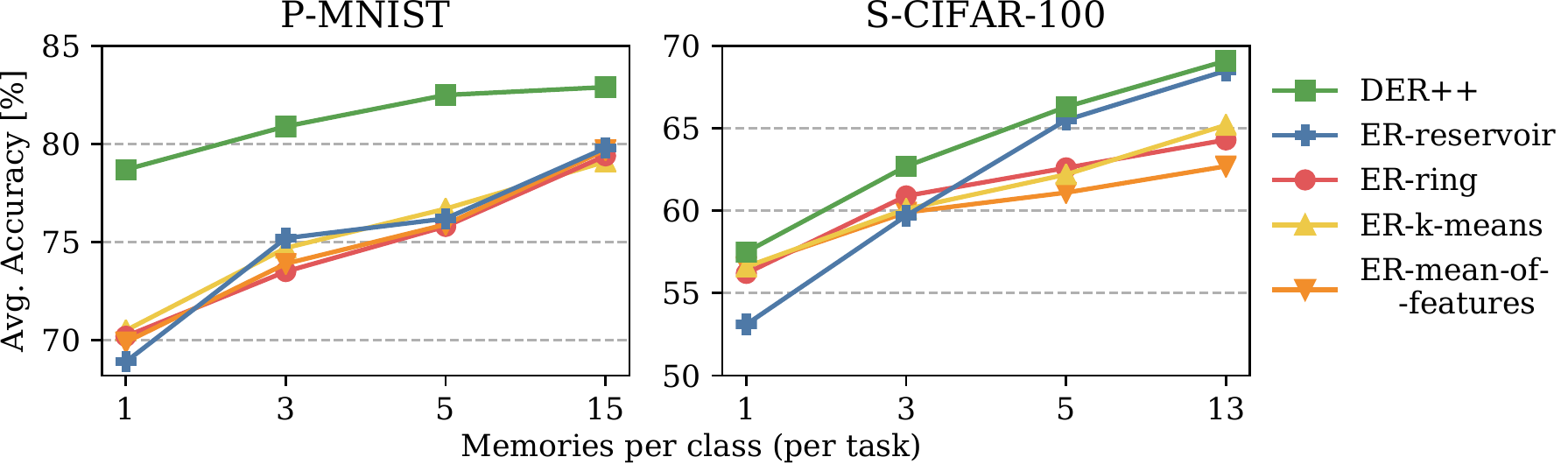}
\caption{A comparison between our proposal (DER++) and the variants of Experience Replay presented in~\cite{chaudhry2019tiny}.}
\label{fig:ciodri}
\end{figure}

In the main paper we already draw a thorough comparison with Experience Replay (ER), showing that DER and DER++ often result in better performance and more remarkable capabilities. It is worth noting that the ER we compared with was equipped with the \textit{reservoir} strategy; therefore, it would be interesting to see whether the same experimental conclusions also hold for other variants of naive replay (\textit{e.g.} ER with ring-buffer). For this reason, Fig.~\ref{fig:ciodri} provides further analysis in the setting of~\cite{chaudhry2019tiny}, which investigates what happens when varying the number of samples that are retained for later replay. Interestingly, while \textit{reservoir} weakens ER when very few past memories are available, it does not bring DER++ to the same flaw. In the low-memory regime, indeed, the probability of leaving a class out from the buffer increases: while ER would not have any chance to retain the knowledge underlying these \quotationmarks{ghost} classes, we conjecture that DER++ could recover this information from the non-argmax outputs of the past predicted distribution.

\subsection{Single-Epoch Setting}
\label{subsec:single_ep}

\begin{table}[H]
    
    \centering
    \begin{tabular}{ccccccc}
    \toprule
     & Buffer & ER & FDR & DER++ & JOINT & JOINT\\ 
    \midrule
    \textbf{\#epochs} & & 1 & 1 & 1 & 1 & 50/100\\ 
    \midrule
    \multirow{3}{*}{\shortstack[c]{Seq.\\CIFAR-10}} & $200$  & $37.64$ & $21.22$ & $\boldsymbol{41.93}$ & & \\ 
     & $500$  & $45.22$ & $21.06$ & $\boldsymbol{48.04}$ & $56.74$ & $92.20$ \\ 
     & $5120$ & $50.28$ & $20.57$ & $\boldsymbol{53.31}$ & & \\ 
    \midrule
    \multirow{3}{*}{\shortstack[c]{Seq.\\Tiny ImageNet}} & $200$   & $5.98$ & $4.87$ &  $\boldsymbol{6.35}$ & & \\ 
    & $500$   & $8.39$ & $4.76$ &  $\boldsymbol{8.65}$ & $19.37$ & $59.99$ \\ 
    & $5120$ & $16.04$ & $4.96$ & $\boldsymbol{16.41}$ & & \\ 
    \bottomrule
    \end{tabular}
    \vspace{0.5em}
    \caption{Single-epoch evaluation setting (Class-IL).}
    \label{tab:singleepoch}
\end{table}

Several Continual Learning works present experiments even on fairly complex datasets (\textit{e.g.:} CIFAR-10, CIFAR-100, Mini ImageNet) in which the model is only trained for one epoch for each task~\cite{aljundi2019gradient, chaudhry2020using, chaudhry2018efficient, lopez2017gradient}. As showing the model each example only once could be deemed closer to real-world CL scenarios, this is a very compelling setting and somewhat close in spirit to the reasons why we focus on General Continual Learning.

However, we see that committing to just one epoch (hence, few gradient steps) makes it difficult to disentangle the effects of catastrophic forgetting (the focus of our work) from those of underfitting. This is especially relevant when dealing with complex datasets and deserves further investigation: for this reason, we conduct a single-epoch experiment on Seq. CIFAR-10 and Seq. Tiny ImageNet. We include in Tab.~\ref{tab:singleepoch} the performance of different rehearsal methods; additionally, we report the results of joint training when limiting the number of epochs to one and, \textit{vice versa}, when such limitation is removed (see last two columns). While the multi-epoch joint training learns to classify with a satisfactory accuracy, the single-epoch counterpart (which is the upper bound to all other methods in this experiment) yields a much lower accuracy and underfits dramatically. In light of this, it is hard to evaluate the merits of other CL methods, whose evaluation is severely undermined by this confounding factor. Although DER++ proves reliable even in this difficult setting, we feel that future CL works should strive for realism by designing experimental settings which are fully in line with the guidelines of GCL~\cite{de2019continual} rather than adopting the single-epoch protocol.

\subsection{Forward and Backward Transfer}
\label{subsec:fwbwt}

In this section, we present additional results for the experiments presented in Sec.~\ref{subsec:cl_exps} and~\ref{subsec:smnist}, reporting \textit{Forward Transfer} (FWT), \textit{Backward Transfer} (BWT)~\cite{lopez2017gradient} and \textit{Forgetting} (FRG)~\cite{chaudhry2018riemannian}. The first one assesses whether a model is capable of improving on unseen tasks w.r.t.\ random guessing, whereas the second and third ones measure the performance degradation in subsequent tasks. Despite their popularity in recent CL works~\cite{chaudhry2018riemannian, chaudhry2020using, de2019continual,lopez2017gradient}, we did not report them in the main paper because we believe that the average accuracy represents an already exhaustive measure of CL performance.

FWT is computed as the difference between the accuracy just before starting training on a given task and the one of the random-initialized network; it is averaged across all tasks. While one can argue that learning to classify unseen classes is desirable, the meaning of such a measure is highly dependent on the setting. Indeed, Class-IL and Task-IL show distinct classes in distinct tasks, which makes transfer impossible. On the contrary, FWT can be relevant for Domain-IL scenarios, provided that the input transformation is not disruptive (as it is the case with Permuted-MNIST). In conclusion, as CL settings sooner or later show all classes to the network, we are primarily interested in the accuracy at the end of the training, not the one before seeing any example.

FRG and BWT compute the difference between the current accuracy and its best value for each task. It is worth noting that any method that restrains the learning of the current task could exhibit high backward transfer but low final accuracy. This is as easy as increasing the weight of the regularization term: this way, the past knowledge is well-preserved but the current task is not learned properly. Moreover, BWT makes the assumption that the highest value of the accuracy on a task is the one yielded at the end of it. This is not always true, as rehearsal-based methods can exploit the memory buffer in a subsequent task, even enhancing their performance on a previous one if they start from low accuracy.

\begin{table}[H]
\centering
{
\setlength{\tabcolsep}{2.0pt}
\resizebox{\textwidth}{!}{
\begin{tabular}{clcccccccc}
\toprule
\multicolumn{8}{c}{\textbf{FORWARD TRANSFER}} \vspace{0.2em}\\
\toprule
 \multirow{2}{*}{\textbf{Buffer}} & \multirow{2}{*}{\textbf{Method}} & \multicolumn{2}{c}{\textbf{S-MNIST}} & \multicolumn{2}{c}{\textbf{S-CIFAR-10}} & \textbf{P-MNIST} & \textbf{R-MNIST}\\
\addlinespace[0.35ex]
 & & \textit{Class-IL} & \textit{Task-IL} & \textit{Class-IL} & \textit{Task-IL} & \textit{Domain-IL} & \textit{Domain-IL}\\
\midrule
\xmark                  & SGD                   & $-11.06$\tiny{$\pm2.90$}                          &  $2.33$\tiny{$\pm4.71$}                           &  $-9.09$\tiny{$\pm0.11$}                          & $-1.46$\tiny{$\pm1.17$}                           &  $0.32$\tiny{$\pm0.85$}                           & $48.94$\tiny{$\pm0.10$} \\
\midrule                            
                        & oEWC                  &  $\boldsymbol{-7.44}$\tiny{$\boldsymbol{\pm4.18}$} & $\boldsymbol{-0.13}$\tiny{$\boldsymbol{\pm8.12}$}& $-12.51$\tiny{$\pm0.02$}                          & $-4.09$\tiny{$\pm7.97$}                           &  $0.69$\tiny{$\pm0.97$}                           & $52.45$\tiny{$\pm8.75$} \\
\multirow{2}{*}{\xmark} & SI                    &  $-9.50$\tiny{$\pm5.27$}                          & $-1.34$\tiny{$\pm5.42$}                           & $-12.64$\tiny{$\pm0.20$}                          & $-2.33$\tiny{$\pm2.29$}                           &  $\boldsymbol{0.71}$\tiny{$\boldsymbol{\pm1.89}$} & $\boldsymbol{53.09}$\tiny{$\boldsymbol{\pm0.73}$} \\
                        & LwF                   & $-12.39$\tiny{$\pm4.06$}                          &  $1.30$\tiny{$\pm5.40$}                           & $\boldsymbol{-10.63}$\tiny{$\boldsymbol{\pm5.12}$} & $\boldsymbol{0.73}$\tiny{$\boldsymbol{\pm4.36}$} & -                                                 & -                       \\
                        & PNN                   & -                                                 & N/A                                               & -                                                 & N/A                                               & -                                                 & -                       \\
\midrule                                                        
                        & ER                    & $-12.12$\tiny{$\pm2.21$}                          & $-0.86$\tiny{$\pm3.24$}                           & $-11.02$\tiny{$\pm2.77$}                          &  $2.10$\tiny{$\pm1.27$}                           &  $1.37$\tiny{$\pm0.48$} & $66.79$\tiny{$\pm0.05$} \\
                        & MER                   & $-11.03$\tiny{$\pm3.40$}                          & $-2.18$\tiny{$\pm3.51$}                           & -                                                 & -                                                 & -                                                 & -                       \\
                        & GEM                   & $-10.26$\tiny{$\pm3.08$}                          & $-0.16$\tiny{$\pm5.89$} &  $-7.50$\tiny{$\pm7.05$}&  $0.13$\tiny{$\pm3.54$}                           &  $0.42$\tiny{$\pm0.35$}                           & $54.06$\tiny{$\pm4.35$}                           \\
                        & A-GEM                 & $\boldsymbol{-10.04}$\tiny{$\boldsymbol{\pm3.11}$} &  $2.39$\tiny{$\pm6.96$}                          & $-11.37$\tiny{$\pm0.08$}                          & $-0.34$\tiny{$\pm0.13$}                           &  $0.83$\tiny{$\pm0.57$}                           & $54.84$\tiny{$\pm10.45$}\\
\multirow{2}{*}{200}    & iCaRL                 & N/A                                               & N/A                                               & N/A                                               & N/A                                               & -                                                 & -                       \\
                        & FDR                   & $-12.06$\tiny{$\pm2.22$}                          & $-0.81$\tiny{$\pm3.89$}                           & $-12.75$\tiny{$\pm0.30$}                          & $-2.42$\tiny{$\pm0.86$}                           & $-1.24$\tiny{$\pm0.06$}                           & $60.71$\tiny{$\pm8.17$} \\
                        & GSS                   & $-11.31$\tiny{$\pm2.58$}                          &  $2.99$\tiny{$\pm6.61$}                           & $-7.08$\tiny{$\pm10.01$}                          & $\boldsymbol{6.17}$\tiny{$\boldsymbol{\pm2.06}$}  &  $0.04$\tiny{$\pm0.85$}                           & $57.28$\tiny{$\pm4.47$} \\
                        & HAL                   & $-11.15$\tiny{$\pm3.56$}                          & $-0.20$\tiny{$\pm3.99$}                           & $-11.94$\tiny{$\pm0.80$}                          & $-0.02$\tiny{$\pm0.10$}                           & $\boldsymbol{1.72}$\tiny{$\boldsymbol{\pm0.08}$}                           & $59.95$\tiny{$\pm3.71$}  \\
                        & \textbf{DER (ours)}   & $-10.16$\tiny{$\pm3.78$}                          &  $\boldsymbol{3.23}$\tiny{$\boldsymbol{\pm5.24}$} & $-11.89$\tiny{$\pm0.88$}                          &  $0.27$\tiny{$\pm7.12$}                           &  $1.23$\tiny{$\pm0.26$}                           & $64.69$\tiny{$\pm2.02$} \\
                        & \textbf{DER++ (ours)} & $-12.42$\tiny{$\pm1.84$}                          & $-2.33$\tiny{$\pm5.69$}                           &  $\boldsymbol{-4.88}$\tiny{$\boldsymbol{\pm6.90}$}&  $2.68$\tiny{$\pm0.11$}                           &  $0.91$\tiny{$\pm0.45$}                           & $\boldsymbol{67.21}$\tiny{$\boldsymbol{\pm2.13}$} \\
\midrule                            
                        & ER                    & $-10.42$\tiny{$\pm3.42$}                          &  $1.02$\tiny{$\pm5.55$}                           &  $-8.42$\tiny{$\pm4.83$}                          & $-3.12$\tiny{$\pm4.02$}                           &  $0.56$\tiny{$\pm2.52$}                           & $65.52$\tiny{$\pm1.56$} \\
                        & MER                   & $-10.59$\tiny{$\pm3.83$}                          &  $0.89$\tiny{$\pm5.03$}                           & -                                                 & -                                                 & -                                                 & -                       \\
                        & GEM                   & $-10.59$\tiny{$\pm3.26$}                          &  $0.11$\tiny{$\pm5.66$}                           & $-12.53$\tiny{$\pm0.65$}                          &  $1.36$\tiny{$\pm3.05$}                           &  $0.17$\tiny{$\pm0.59$}                           & $54.19$\tiny{$\pm2.37$} \\
                        & A-GEM                 &  $-9.74$\tiny{$\pm3.60$}                          &  $1.10$\tiny{$\pm7.30$}                           &  $-6.38$\tiny{$\pm8.64$}                          &  $\boldsymbol{6.36}$\tiny{$\boldsymbol{\pm3.88}$} &  $0.03$\tiny{$\pm1.20$}                           & $52.50$\tiny{$\pm0.51$} \\
\multirow{2}{*}{500}    & iCaRL                 & N/A                                               & N/A                                               & N/A                                               & N/A                                               & -                                                 & -                       \\
                        & FDR                   &  $-9.27$\tiny{$\pm2.80$}                          &  $4.73$\tiny{$\pm5.08$}                           &  $\boldsymbol{-6.23}$\tiny{$\boldsymbol{\pm8.79}$}&  $3.71$\tiny{$\pm2.70$}                           & $-0.32$\tiny{$\pm0.43$}                           & $65.97$\tiny{$\pm1.02$} \\
                        & GSS                   & $-10.16$\tiny{$\pm3.48$}                          &  $0.17$\tiny{$\pm5.32$}                           &  $-7.84$\tiny{$\pm4.43$}                          &  $2.11$\tiny{$\pm3.31$}                           &  $\boldsymbol{0.89}$\tiny{$\boldsymbol{\pm0.94}$} & $58.19$\tiny{$\pm4.42$} \\
                        & HAL                   & $-9.02$\tiny{$\pm5.06$}                           & $0.79$\tiny{$\pm7.26$}                            &  $-7.15$\tiny{$\pm7.57$}                          & $3.06$\tiny{$\pm1.03$}                            & $1.33$\tiny{$\pm0.23$}                           & $64.21$\tiny{$\pm3.16$} \\
                        & \textbf{DER (ours)}   &  $\boldsymbol{-7.96}$\tiny{$\boldsymbol{\pm2.57}$}&  $\boldsymbol{1.17}$\tiny{$\boldsymbol{\pm6.37}$} & $-13.26$\tiny{$\pm1.08$}                          & $-4.52$\tiny{$\pm2.39$}                           &  $0.21$\tiny{$\pm1.21$}                           & $\boldsymbol{72.45}$\tiny{$\boldsymbol{\pm0.14}$} \\
                        & \textbf{DER++ (ours)} & $-10.90$\tiny{$\pm4.88$}                          & $-2.92$\tiny{$\pm5.32$}                           &  $-6.29$\tiny{$\pm8.89$}                          & $-0.31$\tiny{$\pm1.86$}                           & $-0.35$\tiny{$\pm0.01$}                           & $67.05$\tiny{$\pm0.11$} \\
\midrule                            
                        & ER                    & $-10.97$\tiny{$\pm3.70$}                          &  $0.17$\tiny{$\pm3.46$}                           & $-8.45$\tiny{$\pm10.75$}                          & $-1.05$\tiny{$\pm5.87$}                           &  $\boldsymbol{1.46}$\tiny{$\boldsymbol{\pm1.15}$} & $\boldsymbol{73.03}$\tiny{$\boldsymbol{\pm1.59}$} \\
                        & MER                   & $-10.50$\tiny{$\pm3.35$}                          & $-0.33$\tiny{$\pm5.81$}                           & -                                                 & -                                                 & -                                                 & -                       \\
                        & GEM                   &  $-9.51$\tiny{$\pm3.83$}                          & $-0.28$\tiny{$\pm9.16$}                           & $-9.18$\tiny{$\pm4.27$}                           & $-1.24$\tiny{$\pm0.83$}                           &  $1.03$\tiny{$\pm0.89$}                           & $62.06$\tiny{$\pm3.01$} \\
                        & A-GEM                 & $-11.31$\tiny{$\pm3.44$}                          &  $\boldsymbol{1.14}$\tiny{$\boldsymbol{\pm7.08}$} & $-8.01$\tiny{$\pm6.31$}                           & $-3.94$\tiny{$\pm0.82$}                           &  $0.43$\tiny{$\pm0.39$}                           & $51.05$\tiny{$\pm1.34$} \\              
\multirow{2}{*}{5120}   & iCaRL                 & N/A                                               & N/A                                               & N/A                                               & N/A                                               & -                                                 & -                       \\
                        & FDR                   &  $\boldsymbol{-9.25}$\tiny{$\boldsymbol{\pm4.65}$}& $-1.30$\tiny{$\pm5.90$}                           & $-7.69$\tiny{$\pm5.95$}                           & $-0.52$\tiny{$\pm0.54$}                           & $-0.13$\tiny{$\pm0.54$}                           & $72.54$\tiny{$\pm0.35$} \\
                        & GSS                   & $-10.89$\tiny{$\pm3.52$}                          & $-2.19$\tiny{$\pm6.64$}                           & $-9.88$\tiny{$\pm2.21$}                           & $-0.13$\tiny{$\pm5.24$}                                             &  $0.34$\tiny{$\pm1.49$}                           & $63.39$\tiny{$\pm4.55$} \\
                        & HAL                   & $-10.06$\tiny{$\pm4.46$}                          & $0.16$\tiny{$\pm7.43$}                            & $-10.34$\tiny{$\pm3.22$}                          &  $0.32$\tiny{$\pm1.09$}                           & $0.52$\tiny{$\pm0.47$}                          & $66.00$\tiny{$\pm0.09$} \\
                        & \textbf{DER (ours)}   & $-11.59$\tiny{$\pm4.34$}                          & $-2.42$\tiny{$\pm5.22$}                           & $\boldsymbol{-5.98}$\tiny{$\boldsymbol{\pm8.44}$} &  $2.37$\tiny{$\pm3.98$}                           &  $0.32$\tiny{$\pm0.18$}                           & $71.12$\tiny{$\pm0.53$} \\
                        & \textbf{DER++ (ours)} & $-10.71$\tiny{$\pm2.95$}                          &  $0.20$\tiny{$\pm9.44$}                           & $-11.23$\tiny{$\pm2.67$}                          &  $\boldsymbol{4.56}$\tiny{$\boldsymbol{\pm0.02}$} &  $0.06$\tiny{$\pm0.22$}                           & $72.11$\tiny{$\pm1.81$}\\
\bottomrule
\end{tabular}
}
\vspace{0.5em}
\caption{Forward Transfer results for the Experiments of Sec.~\ref{subsec:cl_exps} and~\ref{subsec:smnist}.}
}
\end{table}
\begin{table}[H]
\centering
{
\setlength{\tabcolsep}{2.0pt}
\resizebox{\textwidth}{!}{
\begin{tabular}{clcccccccc}
\toprule
 \multicolumn{8}{c}{\textbf{BACKWARD TRANSFER}} \vspace{0.2em}\\
\toprule
 \multirow{2}{*}{\textbf{Buffer}} & \multirow{2}{*}{\textbf{Method}} & \multicolumn{2}{c}{\textbf{S-MNIST}} & \multicolumn{2}{c}{\textbf{S-CIFAR-10}} & \textbf{P-MNIST} & \textbf{R-MNIST}\\
\addlinespace[0.35ex]
 & & \textit{Class-IL} & \textit{Task-IL} & \textit{Class-IL} & \textit{Task-IL} & \textit{Domain-IL} & \textit{Domain-IL}\\
\midrule
\xmark                  & SGD                   &   $-99.10$\tiny{$\pm0.55$}                            &   $-4.98$\tiny{$\pm2.58$}                         & $-96.39$\tiny{$\pm0.12$}                       & $-46.24$\tiny{$\pm2.12$}                         & $-57.65$\tiny{$\pm4.32$}                      & $-20.34$\tiny{$\pm2.50$} \\
                        \midrule
                        & oEWC                  &   $\boldsymbol{-97.79}$\tiny{$\boldsymbol{\pm1.24}$}  &  $-0.38$\tiny{$\pm0.19$}                          &$\boldsymbol{-91.64}$\tiny{$\boldsymbol{\pm3.07}$}& $-29.13$\tiny{$\pm4.11$}                       & $-36.69$\tiny{$\pm2.34$}                      & $-24.59$\tiny{$\pm5.37$} \\
\multirow{2}{*}{\xmark} & SI                    &   $-98.89$\tiny{$\pm0.86$}                            &   $-3.46$\tiny{$\pm1.69$}                         & $-95.78$\tiny{$\pm0.64$}                       & $-38.76$\tiny{$\pm0.89$}                         & $\boldsymbol{-27.91}$\tiny{$\boldsymbol{\pm0.31}$}& $\boldsymbol{-22.91}$\tiny{$\boldsymbol{\pm0.26}$} \\
                        & LwF                   &   $-99.30$\tiny{$\pm0.11$}                            &   $-6.21$\tiny{$\pm3.67$}                         & $-96.69$\tiny{$\pm0.25$}                       & $-32.56$\tiny{$\pm0.56$}                         & -                                             & -                        \\
                        & PNN                   & -                                                     &  $\boldsymbol{0.00}$\tiny{$\boldsymbol{\pm0.00}$} & -                                              &  $\boldsymbol{0.00}$\tiny{$\boldsymbol{\pm0.00}$}& -                                             & -                        \\
\midrule                                                                                
                        & ER                    &   $-21.36$\tiny{$\pm2.46$}                            &   $-0.82$\tiny{$\pm0.41$}                         & $-61.24$\tiny{$\pm2.62$}                       &  $-7.08$\tiny{$\pm0.64$}                         & $-22.54$\tiny{$\pm0.95$}                      &  $-8.24$\tiny{$\pm1.56$} \\
                        & MER                   &   $-20.38$\tiny{$\pm1.97$}                            &   $-0.81$\tiny{$\pm0.20$}                         & -                                              & -                                                & -                                             & -                       \\
                        & GEM                   &   $-22.32$\tiny{$\pm2.04$}                            &   $-1.14$\tiny{$\pm0.48$}                         & $-82.61$\tiny{$\pm1.60$}                       &  $-9.27$\tiny{$\pm2.07$}                         & $-29.38$\tiny{$\pm2.56$}                      & $-11.51$\tiny{$\pm4.75$} \\
                        & A-GEM                 &   $-66.15$\tiny{$\pm6.84$}                            &  $\boldsymbol{-0.06}$\tiny{$\boldsymbol{\pm2.95}$}& $-95.73$\tiny{$\pm0.20$}                       & $-16.39$\tiny{$\pm0.86$}                         & $-31.69$\tiny{$\pm3.92$}                      & $-19.32$\tiny{$\pm1.17$} \\
\multirow{2}{*}{200}    & iCaRL                 &   $\boldsymbol{-11.73}$\tiny{$\boldsymbol{\pm0.73}$}  &   $-0.23$\tiny{$\pm0.06$}                     & $\boldsymbol{-28.72}$\tiny{$\boldsymbol{\pm0.49}$}&  $\boldsymbol{-1.01}$\tiny{$\boldsymbol{\pm4.15}$}& -                                             & -                        \\
                        & FDR                   &   $-21.15$\tiny{$\pm4.18$}                            &   $-0.50$\tiny{$\pm0.19$}                         & $-86.40$\tiny{$\pm2.67$}                       &  $-7.36$\tiny{$\pm0.03$}                         & $-20.62$\tiny{$\pm0.65$}                      & $-13.31$\tiny{$\pm2.60$} \\
                        & GSS                   &   $-74.10$\tiny{$\pm3.03$}                            &   $-4.29$\tiny{$\pm2.31$}                         & $-75.25$\tiny{$\pm4.07$}                       &  $-8.56$\tiny{$\pm1.78$}                         & $-47.85$\tiny{$\pm1.82$}                      & $-20.19$\tiny{$\pm6.45$} \\
                        & HAL                   &   $-14.54$\tiny{$\pm1.49$}                            &   $-0.48$\tiny{$\pm0.20$}                         & $-69.11$\tiny{$\pm4.21$}                       & $-11.91$\tiny{$\pm0.52$}                         & $-15.24$\tiny{$\pm1.33$}                      & $-11.71$\tiny{$\pm0.26$} \\
                        & \textbf{DER (ours)}   &   $-17.66$\tiny{$\pm2.10$}                            &   $-0.56$\tiny{$\pm0.18$}                         & $-40.76$\tiny{$\pm0.42$}                       &  $-6.21$\tiny{$\pm0.71$}                         & $-13.79$\tiny{$\pm0.80$}                      &  $-5.99$\tiny{$\pm0.46$} \\
                        & \textbf{DER++ (ours)} &   $-16.27$\tiny{$\pm1.73$}                            &   $-0.55$\tiny{$\pm0.37$}                         & $-32.59$\tiny{$\pm2.32$}                       &  $-5.16$\tiny{$\pm0.21$}                         & $\boldsymbol{-11.47}$\tiny{$\boldsymbol{\pm0.33}$}&  $\boldsymbol{-5.27}$\tiny{$\boldsymbol{\pm0.26}$} \\
\midrule                                                                                    
                        & ER                    &   $-15.97$\tiny{$\pm2.46$}                            &   $-0.36$\tiny{$\pm0.20$}                         & $-45.35$\tiny{$\pm0.07$}                       &  $-3.54$\tiny{$\pm0.35$}                         & $-14.90$\tiny{$\pm0.39$}                      &  $-7.52$\tiny{$\pm1.44$} \\
                        & MER                   &   $-11.52$\tiny{$\pm0.56$}                            &   $-0.44$\tiny{$\pm0.17$}                         & -                                              & -                                                & -                                             & -                       \\
                        & GEM                   &   $-15.47$\tiny{$\pm2.03$}                            &   $-0.27$\tiny{$\pm0.98$}                         & $-74.31$\tiny{$\pm4.62$}                       &  $-9.12$\tiny{$\pm0.21$}                         & $-18.76$\tiny{$\pm0.91$}                      &  $-7.19$\tiny{$\pm1.40$} \\
                        & A-GEM                 &  $-65.84$\tiny{$\pm7.24$}                             &   $-0.54$\tiny{$\pm0.20$}                         & $-94.01$\tiny{$\pm1.16$}                       & $-14.26$\tiny{$\pm4.18$}                         & $-28.53$\tiny{$\pm2.01$}                      & $-19.36$\tiny{$\pm3.18$} \\
\multirow{2}{*}{500}    & iCaRL                 &   $-11.84$\tiny{$\pm0.73$}                            &  $\boldsymbol{-0.25}$\tiny{$\boldsymbol{\pm0.09}$}& $-25.71$\tiny{$\pm1.10$}                      &  $\boldsymbol{-1.06}$\tiny{$\boldsymbol{\pm4.21}$}& -                                             & -                        \\
                        & FDR                   &   $-13.90$\tiny{$\pm5.19$}                            &   $-1.27$\tiny{$\pm2.43$}                         & $-85.62$\tiny{$\pm0.36$}                       &  $-4.80$\tiny{$\pm0.30$}                         & $-12.80$\tiny{$\pm1.28$}                      &  $-6.70$\tiny{$\pm1.93$} \\
                        & GSS                   &   $-60.35$\tiny{$\pm6.03$}                            &   $-0.77$\tiny{$\pm0.62$}                         & $-62.88$\tiny{$\pm2.67$}                       &  $-7.73$\tiny{$\pm3.99$}                         & $-23.68$\tiny{$\pm1.35$}                      & $-17.45$\tiny{$\pm9.92$} \\
                        & HAL                   &  $-9.97$\tiny{$\pm1.62$}                              &   $-0.30$\tiny{$\pm0.26$}                         & $-62.21$\tiny{$\pm4.34$}                       &  $-5.41$\tiny{$\pm1.10$}                         & $-11.58$\tiny{$\pm0.49$}                      & $-6.78$\tiny{$\pm0.87$} \\
                        & \textbf{DER (ours)}   &    $-9.58$\tiny{$\pm1.52$}                            &   $-0.39$\tiny{$\pm0.18$}                         & $-26.74$\tiny{$\pm0.15$}                       &  $-4.56$\tiny{$\pm0.45$}                         &  $-8.04$\tiny{$\pm0.42$}                      &  $-3.41$\tiny{$\pm2.18$} \\
                        & \textbf{DER++ (ours)} &    $\boldsymbol{-8.85}$\tiny{$\boldsymbol{\pm1.86}$}  &   $-0.34$\tiny{$\pm0.16$}                         & $\boldsymbol{-22.38}$\tiny{$\boldsymbol{\pm4.41}$}&  $-4.66$\tiny{$\pm1.15$}                      &  $\boldsymbol{-7.62}$\tiny{$\boldsymbol{\pm1.02}$}&  $\boldsymbol{-3.18}$\tiny{$\boldsymbol{\pm0.14}$} \\
\midrule                                                                                    
                        & ER                    &    $-6.07$\tiny{$\pm1.84$}                            &    $0.03$\tiny{$\pm0.36$}                         & $-13.99$\tiny{$\pm1.12$}                       &  $\boldsymbol{0.08}$\tiny{$\boldsymbol{\pm0.06}$}&  $-5.24$\tiny{$\pm0.13$}                      &  $-2.55$\tiny{$\pm0.53$} \\
                        & MER                   &    $\boldsymbol{-3.22}$\tiny{$\boldsymbol{\pm0.33}$}  &    $0.05$\tiny{$\pm0.11$}                         & -                                              & -                                                & -                                             & -                       \\
                        & GEM                   &    $-4.14$\tiny{$\pm1.43$}                            &    $0.16$\tiny{$\pm0.85$}                         & $-75.27$\tiny{$\pm4.41$}                       & $-6.91$\tiny{$\pm2.33$}                          &  $-6.74$\tiny{$\pm0.49$}                      &  $\boldsymbol{-0.06}$\tiny{$\boldsymbol{\pm0.29}$} \\
                        & A-GEM                 &  $-55.04$\tiny{$\pm10.93$}                            &   $\boldsymbol{0.78}$\tiny{$\boldsymbol{\pm4.16}$}& $-84.49$\tiny{$\pm3.08$}                       & $-9.89$\tiny{$\pm0.40$}                          & $-23.73$\tiny{$\pm2.22$}                      & $-17.70$\tiny{$\pm1.28$} \\              
\multirow{2}{*}{5120}   & iCaRL                 &   $-11.64$\tiny{$\pm0.72$}                            &   $-0.22$\tiny{$\pm0.08$}                         & $-24.94$\tiny{$\pm0.14$}                       & $-0.99$\tiny{$\pm1.41$}                          & -                                             & -                        \\
                        & FDR                   &   $-11.58$\tiny{$\pm3.97$}                            &   $-0.87$\tiny{$\pm1.66$}                         & $-96.64$\tiny{$\pm0.19$}                       & $-1.89$\tiny{$\pm0.51$}                          &  $-3.81$\tiny{$\pm0.13$}                      &  $-2.81$\tiny{$\pm0.47$} \\
                        & GSS                   &    $-7.90$\tiny{$\pm1.21$}                            &   $-0.09$\tiny{$\pm0.15$}                         & $-58.11$\tiny{$\pm9.12$}                       & $-6.38$\tiny{$\pm1.71$}                          & $-19.82$\tiny{$\pm1.31$}                      & $-17.05$\tiny{$\pm2.31$} \\
                        & HAL                   &   $-6.55$\tiny{$\pm1.63$}                             &    $0.02$\tiny{$\pm0.20$}                         & $-27.19$\tiny{$\pm7.53$}                       &  $-4.51$\tiny{$\pm0.54$}                         & $-4.27$\tiny{$\pm0.22$}                       & $-2.25$\tiny{$\pm0.01$}  \\
                        & \textbf{DER (ours)}   &    $-4.53$\tiny{$\pm0.83$}                            &   $-0.31$\tiny{$\pm0.08$}                         & $-10.12$\tiny{$\pm0.80$}                       &  $-2.59$\tiny{$\pm0.08$}                         &  $-3.49$\tiny{$\pm0.02$}                      &  $-1.73$\tiny{$\pm0.10$} \\
                        & \textbf{DER++ (ours)} &    $-4.19$\tiny{$\pm1.63$}                            &   $-0.13$\tiny{$\pm0.09$}                         & $\boldsymbol{-6.89}$\tiny{$\boldsymbol{\pm0.50}$}&  $-1.16$\tiny{$\pm0.22$}                       &  $\boldsymbol{-2.93}$\tiny{$\boldsymbol{\pm0.15}$}&  $-1.18$\tiny{$\pm0.53$} \\
\bottomrule
\end{tabular}
}
\vspace{0.5em}
\caption{Backward Transfer results for the Experiments of Sec.~\ref{subsec:cl_exps} and~\ref{subsec:smnist}.}
}
\end{table}
\begin{table}[H]
\centering
{
\setlength{\tabcolsep}{2.0pt}
\resizebox{\textwidth}{!}{
\begin{tabular}{clcccccccc}
\toprule
 \multicolumn{8}{c}{\textbf{FORGETTING}} \vspace{0.2em}\\
\toprule
 \multirow{2}{*}{\textbf{Buffer}} & \multirow{2}{*}{\textbf{Method}} & \multicolumn{2}{c}{\textbf{S-MNIST}} & \multicolumn{2}{c}{\textbf{S-CIFAR-10}} & \textbf{P-MNIST} & \textbf{R-MNIST}\\
\addlinespace[0.35ex]
 & & \textit{Class-IL} & \textit{Task-IL} & \textit{Class-IL} & \textit{Task-IL} & \textit{Domain-IL} & \textit{Domain-IL}\\
\midrule
\xmark                  & SGD                   &  $99.10$\tiny{$\pm0.55$}                                  &   $5.15$\tiny{$\pm2.74$}                                  & $96.39$\tiny{$\pm0.12$}                                           &  $46.24$\tiny{$\pm2.12$}                                          & $57.65$\tiny{$\pm4.32$}                                               & $20.82$\tiny{$\pm2.47$} \\
\midrule                                                                                            
                        & oEWC                  &  $\boldsymbol{97.79}$\tiny{$\boldsymbol{\pm1.24}$}        &   $0.44$\tiny{$\pm0.16$}                                  & $\boldsymbol{91.64}$\tiny{$\boldsymbol{\pm3.07}$}                 &  $29.33$\tiny{$\pm3.84$}                                          & $36.69$\tiny{$\pm2.34$}                                               & $36.44$\tiny{$\pm1.44$} \\
\multirow{2}{*}{\xmark} & SI                    &  $98.89$\tiny{$\pm0.86$}                                  &   $5.15$\tiny{$\pm2.74$}                                  & $95.78$\tiny{$\pm0.64$}                                           &  $38.76$\tiny{$\pm0.89$}                                          & $\boldsymbol{27.91}$\tiny{$\boldsymbol{\pm0.31}$}                     & $\boldsymbol{23.41}$\tiny{$\boldsymbol{\pm0.49}$} \\
                        & LwF                   &  $99.30$\tiny{$\pm0.11$}                                  &   $5.15$\tiny{$\pm2.74$}                                  & $96.69$\tiny{$\pm0.25$}                                           &  $32.56$\tiny{$\pm0.56$}                                          & -                                                                     & -                       \\
                        & PNN                   & -                                                         &   $\boldsymbol{0.00}$\tiny{$\boldsymbol{\pm0.00}$}        & -                                                                 &  $\boldsymbol{0.00}$\tiny{$\boldsymbol{\pm0.00}$} &                                          -                       &                                               -                       \\
\midrule                                                                                            
                        & ER                    &  $21.36$\tiny{$\pm2.46$}                                  &   $0.84$\tiny{$\pm0.41$}                                  & $61.24$\tiny{$\pm2.62$}                                           &   $7.08$\tiny{$\pm0.64$}                                          & $22.54$\tiny{$\pm0.95$}                                               &  $8.87$\tiny{$\pm1.44$} \\
                        & MER                   &  $20.38$\tiny{$\pm1.97$}                                  &   $0.82$\tiny{$\pm0.21$}                                  & -                                                                  & -                                                                & -                                                                     & -                       \\
                        & GEM                   &  $22.32$\tiny{$\pm2.04$}                                  &   $1.19$\tiny{$\pm0.38$}                                  & $82.61$\tiny{$\pm1.60$}                                           &   $9.27$\tiny{$\pm2.07$}                                          & $29.38$\tiny{$\pm2.56$}                                               & $12.97$\tiny{$\pm4.82$} \\
                        & A-GEM                 &  $66.15$\tiny{$\pm6.84$}                                  &   $0.96$\tiny{$\pm0.28$}                                  & $95.73$\tiny{$\pm0.20$}                                           &  $16.39$\tiny{$\pm0.86$}                                          & $31.69$\tiny{$\pm3.92$}                                               & $20.05$\tiny{$\pm1.12$} \\
\multirow{2}{*}{200}    & iCaRL                 &  $\boldsymbol{11.73}$\tiny{$\boldsymbol{\pm0.73}$}        &   $\boldsymbol{0.28}$\tiny{$\boldsymbol{\pm0.08}$}        & $\boldsymbol{28.72}$\tiny{$\boldsymbol{\pm0.49}$}                 &   $\boldsymbol{2.63}$\tiny{$\boldsymbol{\pm3.48}$}                                          & -                                                                     & -                       \\
                        & FDR                   &  $21.15$\tiny{$\pm4.18$}                                  &   $0.52$\tiny{$\pm0.18$}                                  & $86.40$\tiny{$\pm2.67$}                                           &   $7.36$\tiny{$\pm0.03$}                                          & $20.62$\tiny{$\pm0.65$}                                               & $13.66$\tiny{$\pm2.52$} \\
                        & GSS                   &  $74.10$\tiny{$\pm3.03$}                                  &   $4.30$\tiny{$\pm2.31$}                                  & $75.25$\tiny{$\pm4.07$}                                           &   $8.56$\tiny{$\pm1.78$}                                          & $47.85$\tiny{$\pm1.82$}                                               & $20.71$\tiny{$\pm6.50$} \\
                        & HAL                   &  $14.54$\tiny{$\pm1.49$}                                  &   $0.53$\tiny{$\pm0.19$}                                  & $69.11$\tiny{$\pm4.21$}                                           &  $12.26$\tiny{$\pm0.02$}                                          & $79.00$\tiny{$\pm1.17$}                                               & $83.59$\tiny{$\pm0.04$} \\
                        & \textbf{DER (ours)}   &  $17.66$\tiny{$\pm2.10$}                                  &   $0.57$\tiny{$\pm0.18$}                                  & $40.76$\tiny{$\pm0.42$}                                           &   $6.57$\tiny{$\pm0.20$}                                          & $14.00$\tiny{$\pm0.73$}                                               &  $6.53$\tiny{$\pm0.32$} \\
                        & \textbf{DER++ (ours)} &  $16.27$\tiny{$\pm1.73$}                                  &   $0.66$\tiny{$\pm0.28$}                                  & $32.59$\tiny{$\pm2.32$}                                           &   $5.16$\tiny{$\pm0.21$}                                          & $\boldsymbol{11.49}$\tiny{$\boldsymbol{\pm0.31}$}                     &  $\boldsymbol{6.08}$\tiny{$\boldsymbol{\pm0.43}$} \\
\midrule                                                                                            
                        & ER                    &  $15.97$\tiny{$\pm2.46$}                                  &   $0.39$\tiny{$\pm0.20$}                                  & $45.35$\tiny{$\pm0.07$}                                           &   $3.54$\tiny{$\pm0.35$}                                          & $14.90$\tiny{$\pm0.39$}                                               &  $8.02$\tiny{$\pm1.56$} \\
                        & MER                   &  $11.52$\tiny{$\pm0.56$}                                  &   $0.45$\tiny{$\pm0.17$}                                  & -                                                                  & -                                                                & -                                                                     & -                       \\
                        & GEM                   &  $15.57$\tiny{$\pm1.77$}                                  &   $0.54$\tiny{$\pm0.15$}                                  & $74.31$\tiny{$\pm4.62$}                                           &   $9.12$\tiny{$\pm0.21$}                                          & $18.76$\tiny{$\pm0.91$}                                               &  $8.79$\tiny{$\pm1.44$} \\
                        & A-GEM                 &  $65.84$\tiny{$\pm7.24$}                                  &   $0.64$\tiny{$\pm0.20$}                                  & $94.01$\tiny{$\pm1.16$}                                           &  $14.26$\tiny{$\pm4.18$}                                          & $28.53$\tiny{$\pm2.01$}                                               & $19.70$\tiny{$\pm3.14$} \\
\multirow{2}{*}{500}    & iCaRL                 &  $11.84$\tiny{$\pm0.73$}                                  &   $\boldsymbol{0.30}$\tiny{$\boldsymbol{\pm0.09}$}        & $25.71$\tiny{$\pm1.10$}                                           &   $\boldsymbol{2.66}$\tiny{$\boldsymbol{\pm2.47}$}                & -                                                                     & -                       \\
                        & FDR                   &  $13.90$\tiny{$\pm5.19$}                                  &   $1.35$\tiny{$\pm2.40$}                                  & $85.62$\tiny{$\pm0.36$}                                           &   $4.80$\tiny{$\pm0.00$}                                          & $12.80$\tiny{$\pm1.28$}                                               &  $7.21$\tiny{$\pm1.89$} \\
                        & GSS                   &  $60.35$\tiny{$\pm6.03$}                                  &   $0.89$\tiny{$\pm0.40$}                                  &  $62.88$\tiny{$\pm2.67$}                                          &   $7.73$\tiny{$\pm3.99$}                                          & $23.68$\tiny{$\pm1.35$}                                               & $18.05$\tiny{$\pm9.89$} \\
                        & HAL                   &   $9.97$\tiny{$\pm1.62$}                                  &   $0.35$\tiny{$\pm0.21$}                                  & $62.21$\tiny{$\pm4.34$}                                           &   $5.41$\tiny{$\pm1.10$}                                          & $82.53$\tiny{$\pm0.36$}                                               & $88.53$\tiny{$\pm0.77$}  \\
                        & \textbf{DER (ours)}   &   $9.58$\tiny{$\pm1.52$}                                  &   $0.45$\tiny{$\pm0.13$}                                  & $26.74$\tiny{$\pm0.15$}                                           &   $4.56$\tiny{$\pm0.45$}                                          &  $8.07$\tiny{$\pm0.43$}                                               &  $3.96$\tiny{$\pm2.08$} \\
                        & \textbf{DER++ (ours)} &   $\boldsymbol{8.85}$\tiny{$\boldsymbol{\pm1.86}$}        &   $0.35$\tiny{$\pm0.15$}                                  & $\boldsymbol{22.38}$\tiny{$\boldsymbol{\pm4.41}$}                 &   $4.66$\tiny{$\pm1.15$}                                          &  $\boldsymbol{7.67}$\tiny{$\boldsymbol{\pm1.05}$}                     &  $\boldsymbol{3.57}$\tiny{$\boldsymbol{\pm0.09}$} \\
\midrule                                                                                            
                        & ER                    &   $6.08$\tiny{$\pm1.84$}                                  &   $0.25$\tiny{$\pm0.23$}                                  & $13.99$\tiny{$\pm1.12$}                                           &   $0.27$\tiny{$\pm0.06$}                                          &  $5.24$\tiny{$\pm0.13$}                                               &  $3.10$\tiny{$\pm0.42$} \\
                        & MER                   &   $\boldsymbol{3.22}$\tiny{$\boldsymbol{\pm0.33}$}        &   $\boldsymbol{0.07}$\tiny{$\boldsymbol{\pm0.06}$}        & -                                                                  & -                                                                & -                                                                     & -                       \\
                        & GEM                   &   $4.30$\tiny{$\pm1.16$}                                  &   $0.16$\tiny{$\pm0.09$}                                  & $75.27$\tiny{$\pm4.41$}                                           &  $6.91$\tiny{$\pm2.33$}                                          &  $6.74$\tiny{$\pm0.49$}                                               &  $2.49$\tiny{$\pm0.17$} \\
                        & A-GEM                 & $55.10$\tiny{$\pm10.79$}                                  &   $0.63$\tiny{$\pm0.21$}                                  & $84.49$\tiny{$\pm3.08$}                                           &  $11.36$\tiny{$\pm1.68$}                                          & $23.74$\tiny{$\pm2.23$}                                               & $18.10$\tiny{$\pm1.44$} \\              
\multirow{2}{*}{5120}   & iCaRL                 &  $11.64$\tiny{$\pm0.72$}                                  &   $0.26$\tiny{$\pm0.06$}                                  & $24.94$\tiny{$\pm0.14$}                                           &   $1.59$\tiny{$\pm0.57$}                                          & -                                                                     & -                       \\
                        & FDR                   &  $11.58$\tiny{$\pm3.97$}                                  &   $0.95$\tiny{$\pm1.61$}                                  & $96.64$\tiny{$\pm0.19$}                                           &   $1.93$\tiny{$\pm0.48$}                                          &  $3.82$\tiny{$\pm0.12$}                                               &  $3.31$\tiny{$\pm0.56$} \\
                        & GSS                   &   $7.90$\tiny{$\pm1.21$}                                  &   $0.18$\tiny{$\pm0.11$}                                  & $58.11$\tiny{$\pm9.12$}                                            & $7.71$\tiny{$\pm2.31$}                                             & $89.76$\tiny{$\pm0.39$}                                                &   $92.66$\tiny{$\pm0.02$} \\
                        & HAL                   &   $6.55$\tiny{$\pm1.63$}                                  &   $0.13$\tiny{$\pm0.07$}                                  & $27.19$\tiny{$\pm7.53$}                                           &   $5.21$\tiny{$\pm0.50$}                                          & $19.97$\tiny{$\pm1.33$}                                               & $17.62$\tiny{$\pm2.33$} \\
                        & \textbf{DER (ours)}   &   $4.53$\tiny{$\pm0.83$}                                  &   $0.32$\tiny{$\pm0.08$}                                  & $10.12$\tiny{$\pm0.80$}                                           &   $2.59$\tiny{$\pm0.08$}                                          &  $3.51$\tiny{$\pm0.03$}                                               &  $2.17$\tiny{$\pm0.11$} \\
                        & \textbf{DER++ (ours)} &   $4.19$\tiny{$\pm1.63$}                                  &   $0.23$\tiny{$\pm0.06$}                                  &  $\boldsymbol{7.27}$\tiny{$\boldsymbol{\pm0.84}$}                 &   $\boldsymbol{1.18}$\tiny{$\boldsymbol{\pm0.19}$}                &  $\boldsymbol{2.96}$\tiny{$\boldsymbol{\pm0.14}$}                     &  $\boldsymbol{1.62}$\tiny{$\boldsymbol{\pm0.50}$} \\
\bottomrule
\end{tabular}
}
\caption{Forgetting results for the Experiments of Sec.~\ref{subsec:cl_exps} and~\ref{subsec:smnist}.}
}
\end{table}

\section{Hyperparameter Search}
\subsection{Best values}
In Table~\ref{tab:hyperparams}, we show the best hyperparameter combination that we chose for each method for the experiments in the main paper, according to the criteria outlined in Section~\ref{subsec:cl_eval_prot}. We denote the learning rate with \textit{lr}, the batch size with \textit{bs} and the minibatch size (i.e. the size of the batches drawn from the buffer in rehearsal-based methods) with \textit{mbs}, while other symbols refer to the respective methods. We hold \textit{batch size} and \textit{minibatch size} out of the hyperparameter search space for all Continual Learning benchmarks. Their values are fixed as follows: Sequential MNIST: $10$; Sequential CIFAR-10, Sequential Tiny ImageNet: $32$; Permuted MNIST, Rotated MNIST: $128$.

Conversely, \textit{batch size} and \textit{minibatch size} belong to the hyperparameter search space for experiments on the novel MNIST-360 dataset. It must be noted that MER does not depend on \textit{batch size}, as it internally always adopts a single-example forward pass.
\setlength\cellspacetoplimit{3.5pt}
\setlength\cellspacebottomlimit{1.5pt}

\begin{table}[H]
\centering\begin{tabular}{| Sl | Sc | Sl | Sc | Sl |}
\hline
\textit{Method} & \textit{Buffer} & \textbf{Permuted MNIST} & \textit{Buffer} & \textbf{Rotated MNIST}\\ 
\hline
SGD     & \xmark & \textit{lr:} $0.2$ & \xmark & \textit{lr:} $0.2$ \\
oEWC    & \xmark & \textit{lr:} $0.1$ ~~ \textit{$\lambda$:} $0.7$ ~~ \textit{$\gamma$:} $1.0$ & \xmark & \textit{lr:} $0.1$ ~~ \textit{$\lambda$:} $0.7$ ~~ \textit{$\gamma$:} $1.0$ \\
SI      & \xmark & \textit{lr:} $0.1$ ~~ \textit{c:} $0.5$ ~~ \textit{$\xi$:} $1.0$ & \xmark & \textit{lr:} $0.1$ ~~ \textit{c:} $1.0$ ~~ \textit{$\xi$:} $1.0$ \\
ER      & 200    & \textit{lr:} $0.2$       & 200  &\textit{lr:} $0.2$   \\
        & 500    & \textit{lr:} $0.2$       & 500  &\textit{lr:} $0.2$    \\
        & 5120   & \textit{lr:} $0.2$       & 5120 &\textit{lr:} $0.2$   \\
GEM     & 200    & \textit{lr:} $0.1$ ~~ \textit{$\gamma$:} $0.5$ & 200  &\textit{lr:} $0.01$ ~~  \textit{$\gamma$:} $0.5$ \\
        & 500    & \textit{lr:} $0.1$ ~~ \textit{$\gamma$:} $0.5$ & 500  &\textit{lr:} $0.01$ ~~ \textit{$\gamma$:} $0.5$ \\
        & 5120   & \textit{lr:} $0.1$ ~~ \textit{$\gamma$:} $0.5$ & 5120 &\textit{lr:} $0.01$ ~~ \textit{$\gamma$:} $0.5$ \\
A-GEM   & 200    & \textit{lr:} $0.1$       & 200  &\textit{lr:} $0.1$   \\
        & 500    & \textit{lr:} $0.1$       & 500  &\textit{lr:} $0.3$   \\
        & 5120   & \textit{lr:} $0.1$       & 5120 &\textit{lr:} $0.3$   \\
FDR     & 200    & \textit{lr:} $0.1$  ~~ \textit{$\alpha$:} $1.0$ & 200  & \textit{lr:} $0.1$  ~~ \textit{$\alpha$:} $1.0$ \\
        & 500    & \textit{lr:} $0.1$  ~~ \textit{$\alpha$:} $0.3$ & 500  & \textit{lr:} $0.2$  ~~ \textit{$\alpha$:} $0.3$ \\
        & 5120   & \textit{lr:} $0.1$  ~~ \textit{$\alpha$:} $1.0$ & 5120 & \textit{lr:} $0.2$  ~~ \textit{$\alpha$:} $1.0$ \\
GSS     & 200 & \textit{lr:} $0.2$  ~~ \textit{$gmbs$:} $128$ ~~ \textit{$nb$:} $1$ & 200 & \textit{lr:} $0.2$  ~~ \textit{$gmbs$:} $128$ ~~ \textit{$nb$:} $1$ \\
        & 500    & \textit{lr:} $0.1$  ~~ \textit{$gmbs$:} $10$ ~~ \textit{$nb$:} $1$ & 500 & \textit{lr:} $0.2$  ~~ \textit{$gmbs$:} $128$ ~~ \textit{$nb$:} $1$ \\
        & 5120   & \textit{lr:} $0.03$  ~~ \textit{$gmbs$:} $10$ ~~ \textit{$nb$:} $1$ & 5120 & \textit{lr:} $0.2$  ~~ \textit{$gmbs$:} $128$ ~~ \textit{$nb$:} $1$ \\
HAL     & 200    & \textit{lr:} $0.1$  ~ \textit{$\lambda$:} $0.1$ ~ \textit{$\beta$:} $0.5$ ~ \textit{$\gamma$:} $0.1$  & 200  & \textit{lr:} $0.1$  ~ \textit{$\lambda$:} $0.2$~~ \textit{$\beta$:} $0.5$ ~ \textit{$\gamma$:} $0.1$  \\
        & 500    & \textit{lr:} $0.1$  ~ \textit{$\lambda$:} $0.1$ ~ \textit{$\beta$:} $0.3$ ~ \textit{$\gamma$:} $0.1$   & 500  & \textit{lr:} $0.1$  ~ \textit{$\lambda$:} $0.1$ ~ \textit{$\beta$:} $0.5$ ~ \textit{$\gamma$:} $0.1$  \\
        & 5120   & \textit{lr:} $0.1$  ~ \textit{$\lambda$:} $0.1$ ~ \textit{$\beta$:} $0.5$ ~ \textit{$\gamma$:} $0.1$   & 5120 & \textit{lr:} $0.1$  ~ \textit{$\lambda$:} $0.1$ ~ \textit{$\beta$:} $0.3$ ~ \textit{$\gamma$:} $0.1$  \\
DER     & 200    & \textit{lr:} $0.2$  ~~ \textit{$\alpha$:} $1.0$ & 200  & \textit{lr:} $0.2$  ~~ \textit{$\alpha$:} $1.0$ \\
        & 500    & \textit{lr:} $0.2$  ~~ \textit{$\alpha$:} $1.0$ & 500  & \textit{lr:} $0.2$  ~~ \textit{$\alpha$:} $0.5$ \\
        & 5120   & \textit{lr:} $0.2$  ~~ \textit{$\alpha$:} $0.5$ & 5120 & \textit{lr:} $0.2$  ~~ \textit{$\alpha$:} $0.5$ \\
DER++   & 200    & \textit{lr:} $0.1$  ~~ \textit{$\alpha$:} $1.0$ ~~ \textit{$\beta$:} $1.0$ & 200  & \textit{lr:} $0.1$  ~~ \textit{$\alpha$:} $1.0$ ~~ \textit{$\beta$:} $0.5$ \\
        & 500    & \textit{lr:} $0.2$  ~~ \textit{$\alpha$:} $1.0$ ~~ \textit{$\beta$:} $0.5$ & 500  & \textit{lr:} $0.2$  ~~ \textit{$\alpha$:} $0.5$ ~~ \textit{$\beta$:} $1.0$ \\
        & 5120   & \textit{lr:} $0.2$  ~~ \textit{$\alpha$:} $0.5$ ~~ \textit{$\beta$:} $1.0$ & 5120 & \textit{lr:} $0.2$  ~~ \textit{$\alpha$:} $0.5$ ~~ \textit{$\beta$:} $0.5$ \\
        & & & & \\
\hline
\textit{Method} & \textit{Buffer} & \textbf{Sequential MNIST} & \textit{Buffer} & \textbf{Sequential CIFAR-10}  \\
\hline
SGD & \xmark & \textit{lr:} $0.03$  ~~ & \xmark & \textit{lr:} $0.1$ \\
oEWC    & \xmark & \textit{lr:} $0.03$ ~~ \textit{$\lambda$:} $90$ ~~ \textit{$\gamma$:} $1.0$ & \xmark & \textit{lr:} $0.03$ ~~ \textit{$\lambda$:} $10$ ~~ \textit{$\gamma$:} $1.0$ \\
SI      & \xmark & \textit{lr:} $0.1$ ~~ \textit{c:} $1.0$ ~~ \textit{$\xi$:} $0.9$ & \xmark & \textit{lr:} $0.03$ ~~ \textit{c:} $0.5$ ~~ \textit{$\xi$:} $1.0$ \\
LwF     & \xmark & \textit{lr:} $0.03$ ~ \textit{$\alpha$: $1$} ~ \textit{T:} $2.0$ ~ \textit{wd:} $0.0005$ & \xmark & \textit{lr:} $0.03$ ~~ \textit{$\alpha$: $0.5$} ~~ \textit{T:} $2.0$ \\
PNN     & \xmark & \textit{lr:} $0.1$  ~~ & \xmark & \textit{lr:} $0.03$ \\
ER      & 200    & \textit{lr:} $0.01$   & 200  &\textit{lr:} $0.1$  \\
        & 500    & \textit{lr:} $0.1$    & 500  &\textit{lr:} $0.1$  \\
        & 5120   & \textit{lr:} $0.1$   & 5120 &\textit{lr:} $0.1$  \\
        & & & & \\
MER     & 200    & \textit{lr:} $0.1$   ~ \textit{$\beta$:} $1$ ~ \textit{$\gamma$:} $1$ ~ \textit{nb:} $1$ ~ \textit{bs:} $1$ & & \\
        & 500    & \textit{lr:} $0.1$  ~ \textit{$\beta$:} $1$ ~ \textit{$\gamma$:} $1$ ~ \textit{nb:} $1$ ~ \textit{bs:} $1$ & & \\
        & 5120   & \textit{lr:} $0.03$   ~ \textit{$\beta$:} $1$ ~ \textit{$\gamma$:} $1$ ~ \textit{nb:} $1$ ~ \textit{bs:} $1$ & & \\
GEM     & 200    & \textit{lr:} $0.01$  ~~  \textit{$\gamma$:} $1.0$ & 200  &\textit{lr:} $0.03$ ~~ \textit{$\gamma$:} $0.5$ \\
        & 500    & \textit{lr:} $0.03$  ~~  \textit{$\gamma$:} $0.5$ & 500  &\textit{lr:} $0.03$ ~~ \textit{$\gamma$:} $0.5$ \\
        & 5120   & \textit{lr:} $0.1$  ~~  \textit{$\gamma$:} $1.0$ & 5120 &\textit{lr:} $0.03$ ~~ \textit{$\gamma$:} $0.5$ \\
\hline
\end{tabular}
\end{table}

\begin{table}[H]
\centering\begin{tabular}{| Sl | Sc | Sl | Sc | Sl |}
\hline
\textit{Method} & \textit{Buffer} & \textbf{Sequential MNIST} & \textit{Buffer} & \textbf{Sequential CIFAR-10} \\
\hline
A-GEM   & 200    & \textit{lr:} $0.1$   & 200  & \textit{lr:} $0.03$  \\
        & 500    & \textit{lr:} $0.1$   & 500  & \textit{lr:} $0.03$  \\
        & 5120   & \textit{lr:} $0.1$   & 5120 & \textit{lr:} $0.03$  \\
iCaRL   & 200    & \textit{lr:} $0.1$ ~ \textit{wd:} $0$ & 200  & \textit{lr:} $0.1$ ~~ \textit{wd:} $10^{-5}$ \\
        & 500    & \textit{lr:} $0.1$ ~~ \textit{wd:} $0$ & 500  & \textit{lr:} $0.1$ ~~ \textit{wd:}$10^{-5}$ \\
        & 5120   & \textit{lr:} $0.1$ ~~ \textit{wd:} $0$ & 5120 & \textit{lr:} $0.03$ ~~ \textit{wd:}$10^{-5}$ \\
FDR     & 200    & \textit{lr:} $0.03$   ~~ \textit{$\alpha$:} $0.5$ & 200  & \textit{lr:} $0.03$  ~~ \textit{$\alpha$:} $0.3$ \\
        & 500    & \textit{lr:} $0.1$   ~~ \textit{$\alpha$:} $0.2$ & 500  & \textit{lr:} $0.03$  ~~ \textit{$\alpha$:} $1.0$ \\
        & 5120   & \textit{lr:} $0.1$   ~~ \textit{$\alpha$:} $0.2$ & 5120 & \textit{lr:} $0.03$  ~~ \textit{$\alpha$:} $0.3$ \\
GSS     & 200    & \textit{lr:} $0.1$  ~~ \textit{$gmbs$:} $10$ ~~ \textit{$nb$:} $1$ & 200  & \textit{lr:} $0.03$ ~~ \textit{$gmbs$:} $32$ ~~ \textit{$nb$:} $1$\\
        & 500    & \textit{lr:} $0.1$  ~~ \textit{$gmbs$:} $10$ ~~ \textit{$nb$:} $1$ & 500  & \textit{lr:} $0.03$  ~~ \textit{$gmbs$:} $32$ ~~ \textit{$nb$:} $1$\\
        & 5120   & \textit{lr:} $0.1$  ~~ \textit{$gmbs$:} $10$ ~~ \textit{$nb$:} $1$ & 5120 & \textit{lr:} $0.03$  ~~ \textit{$gmbs$:} $32$ ~~ \textit{$nb$:} $1$\\
HAL     & 200    & \textit{lr:} $0.1$  ~ \textit{$\lambda$:} $0.1$ ~ \textit{$\beta$:} $0.7$ ~ \textit{$\gamma$:} $0.5$  & 200  & \textit{lr:} $0.03$ ~~            \textit{$\lambda$:} $0.2$ ~~ \textit{$\beta$:} $0.5$ \\
        & & & & \textit{$\gamma$:} $0.1$ \\
        & 500    & \textit{lr:} $0.1$  ~ \textit{$\lambda$:} $0.1$ ~ \textit{$\beta$:} $0.2$ ~ \textit{$\gamma$:} $0.5$  & 500  & \textit{lr:} $0.03$  ~~ \textit{$\lambda$:} $0.1$ ~~ \textit{$\beta$:} $0.3$ \\
        & & & & \textit{$\gamma$:} $0.1$ \\
        & 5120   & \textit{lr:} $0.1$  ~ \textit{$\lambda$:} $0.1$ ~ \textit{$\beta$:} $0.7$ ~ \textit{$\gamma$:} $0.5$ & 5120 & \textit{lr:} $0.03$ ~~ \textit{$\lambda$:} $0.1$ ~~ \textit{$\beta$:} $0.3$ \\
        & & & & \textit{$\gamma$:} $0.1$ \\
DER     & 200    & \textit{lr:} $0.03$   ~~ \textit{$\alpha$:} $0.2$ & 200  & \textit{lr:} $0.03$  ~~ \textit{$\alpha$:} $0.3$ \\
        & 500    & \textit{lr:} $0.03$   ~~ \textit{$\alpha$:} $1.0$ & 500  & \textit{lr:} $0.03$  ~~ \textit{$\alpha$:} $0.3$ \\
        & 5120   & \textit{lr:} $0.1$   ~~ \textit{$\alpha$:} $0.5$ & 5120 & \textit{lr:} $0.03$  ~~ \textit{$\alpha$:} $0.3$ \\
DER++   & 200    & \textit{lr:} $0.03$  ~~ \textit{$\alpha$:} $0.2$ ~~ \textit{$\beta$:} $1.0$ & 200 & \textit{lr:} $0.03$  ~~ \textit{$\alpha$:} $0.1$ ~~ \textit{$\beta$:} $0.5$ \\
        & 500    & \textit{lr:} $0.03$    ~~ \textit{$\alpha$:} $1.0$ ~~ \textit{$\beta$:} $0.5$ & 500 & \textit{lr:} $0.03$  ~~ \textit{$\alpha$:} $0.2$ ~~ \textit{$\beta$:} $0.5$ \\
        & 5120   & \textit{lr:} $0.1$    ~~ \textit{$\alpha$:} $0.2$ ~~ \textit{$\beta$:} $0.5$ & 5120 & \textit{lr:} $0.03$  ~~ \textit{$\alpha$:} $0.1$ ~~ \textit{$\beta$:} $1.0$ \\
        & & & & \\
\hline
\textit{Method} & \textit{Buffer} & \textbf{Sequential Tiny ImageNet} & \textit{Buffer} & \textbf{MNIST-360} \\
\hline
SGD & \xmark & \textit{lr:} $0.03$ & \xmark & \textit{lr:} $0.1$ ~~ \textit{bs:} $4$ \\
oEWC    & \xmark & \textit{lr:} $0.03$ ~~ \textit{$\lambda$:} $25$ ~~ \textit{$\gamma$:} $1.0$ & & \\
SI      & \xmark & \textit{lr:} $0.03$ ~~ \textit{c:} $0.5$ ~~ \textit{$\xi$:} $1.0$ & & \\
LwF     & \xmark & \textit{lr:} $0.01$ ~~ \textit{$\alpha$: $1.0$} ~~ \textit{T:} $2.0$ & & \\
PNN     & \xmark & \textit{lr:} $0.03$ & & \\
ER      & 200    & \textit{lr:} $0.1$  & 200    & \textit{lr:} $0.2$ ~~ \textit{bs:} $1$ ~~ \textit{mbs:} $16$ \\
        & 500    & \textit{lr:} $0.03$   & 500    & \textit{lr:} $0.2$ ~~ \textit{bs:} $1$ ~~ \textit{mbs:} $16$ \\
        & 5120   & \textit{lr:} $0.1$  & 1000   & \textit{lr:} $0.2$ ~~ \textit{bs:} $4$ ~~ \textit{mbs:} $16$ \\
MER     & & & 200  & \textit{lr:} $0.2$  ~~  \textit{mbs:} $128$ ~~ \textit{$\beta$:} $1$ \\
        & & & & \textit{$\gamma$:} $1$ ~~ \textit{nb:} $3$ \\
        & & & 500  & \textit{lr:} $0.1$  ~~  \textit{mbs:} $128$ ~~ \textit{$\beta$:} $1$ \\
        & & & & \textit{$\gamma$:} $1$ ~~ \textit{nb:} $3$\\
        & & & 1000 & \textit{lr:} $0.2$ ~~  \textit{mbs:} $128$ ~~ \textit{$\beta$:} $1$ \\
        & & & & \textit{$\gamma$:} $1$ ~~ \textit{nb:} $3$\\
A-GEM  & 200    & \textit{lr:} $0.01$  &
200    & \textit{lr:} $0.1$ ~~ \textit{bs:} $16$ ~~ \textit{mbs:} $128$ \\
        & 500    & \textit{lr:} $0.01$  &
500    & \textit{lr:} $0.1$ ~~ \textit{bs:} $16$ ~~ \textit{mbs:} $128$ \\
        & 5120   & \textit{lr:} $0.01$  &
1000    & \textit{lr:} $0.1$ ~~ \textit{bs:} $4$ ~~ \textit{mbs:} $128$\\
iCaRL   & 200    & \textit{lr:} $0.03$ ~~ \textit{wd:} $10^{-5}$ & & \\
        & 500    & \textit{lr:} $0.03$ ~~ \textit{wd:} $10^{-5}$ & & \\
        & 5120   & \textit{lr:} $0.03$ ~~ \textit{wd:} $10^{-5}$ & & \\
\hline
\end{tabular}
\end{table}

\begin{table}[H]
\centering\begin{tabular}{| Sl | Sc | Sl | Sc | Sl |}
\hline
\textit{Method} & \textit{Buffer} & \textbf{Sequential Tiny Imagenet} & \textit{Buffer} & \textbf{MNIST-360} \\
\hline
FDR      & 200   & \textit{lr:} $0.03$ ~~ \textit{$\alpha$:} $0.3$ & & \\
        & 500    & \textit{lr:} $0.03$  ~~ \textit{$\alpha$:} $1.0$ & &\\
        & 5120   & \textit{lr:} $0.03$  ~~ \textit{$\alpha$:} $0.3$ & & \\
GSS      & & & 200    & \textit{lr:} $0.2$ ~~ \textit{bs:} $1$ ~~ \textit{mbs:} $16$ \\
        & & & 500    & \textit{lr:} $0.2$ ~~ \textit{bs:} $1$ ~~ \textit{mbs:} $16$ \\
        & & & 1000   & \textit{lr:} $0.2$ ~~ \textit{bs:} $4$ ~~ \textit{mbs:} $16$ \\
DER      & 200   & \textit{lr:} $0.03$ ~~ \textit{$\alpha$:} $0.1$ & 200  & \textit{lr:} $0.1$ ~~ \textit{bs:} $16$ \\
        & & & & \textit{mbs:} $64$ ~~ \textit{$\alpha$:} $0.5$\\
        & 500    & \textit{lr:} $0.03$  ~~ \textit{$\alpha$:} $0.1$ & 500  & \textit{lr:} $0.2$ ~~ \textit{bs:} $16$ \\
        & & & & \textit{mbs:} $16$ ~~ \textit{$\alpha$:} $0.5$\\
        & 5120   & \textit{lr:} $0.03$  ~~ \textit{$\alpha$:} $0.1$ & 1000 & \textit{lr:} $0.1$ ~~ \textit{bs:} $8$ \\
        & & & & \textit{mbs:} $16$ ~~ \textit{$\alpha$:} $0.5$\\
DER++   & 200    & \textit{lr:} $0.03$  ~~ \textit{$\alpha$:} $0.1$ ~~ \textit{$\beta$:} $1.0$ & 200    & \textit{lr:} $0.2$ ~ \textit{bs:} $16$ \\
        & & & & \textit{mbs:} $16$ ~~ \textit{$\alpha$:} $0.5$ ~~ \textit{$\beta$:} $1.0$ \\
        & 500    & \textit{lr:} $0.03$  ~~ \textit{$\alpha$:} $0.2$ ~~ \textit{$\beta$:} $0.5$ & 500    & \textit{lr:} $0.2$ ~~ \textit{bs:} $16$ \\
        & & & & \textit{mbs:} $16$ ~ \textit{$\alpha$:} $0.5$ ~~ \textit{$\beta$:} $1.0$ \\
        & 5120   & \textit{lr:} $0.03$  ~~ \textit{$\alpha$:} $0.1$ ~~ \textit{$\beta$:} $0.5$ & 1000   & \textit{lr:} $0.2$ ~~ \textit{bs:} $16$ \\
        & & & & \textit{mbs:} $128$ ~~ \textit{$\alpha$:} $0.2$ ~~ \textit{$\beta$:} $1.0$ \\
\hline
\end{tabular}
\vspace{0.8em}
\caption{Hyperparameters selected for our experiments.}
\label{tab:hyperparams}
\end{table}

\subsection{All values}
In the following, we provide a list of all the parameter combinations that were considered (Table~\ref{tab:hyperparam_space}). Note that the same parameters are searched for all examined buffer sizes.

{
\begin{table}[H]
    \begin{multicols}{3}
        \setlength{\tabcolsep}{2.5pt}
        \scriptsize
        \begin{tabular}{ccc}
\toprule
\multicolumn{3}{c}{\textbf{Permuted MNIST}}\\
\textit{Method} &  \textit{Par} &
\textit{Values} \\
\midrule
SGD/JOINT  & \textit{lr} & [0.03, 0.1, 0.2] \\
\midrule
oEWC & \textit{lr} & [0.1, 0.01] \\
& 	 \textit{$\lambda$} & [0.1, 1, 10, 30, 90,100] \\
& 	 \textit{$\gamma$} & [0.9, 1.0] \\
\midrule
SI  & \textit{lr} & [0.01, 0.1] \\
& 	 \textit{c} & [0.5, 1.0] \\
& 	 \textit{$\xi$} & [0.9, 1.0] \\
\midrule
ER & \textit{lr} & [0.03, 0.1, 0.2] \\
\midrule
GEM & \textit{lr} & [0.01, 0.1, 0.3] \\
& \textit{$\gamma$} & [0.1, 0.5, 1]\\
\midrule
A-GEM  & \textit{lr} & [0.01, 0.1, 0.3] \\
\midrule
GSS  & \textit{lr} & [0.03, 0.1, 0.2] \\
& 	 \textit{gmbs} & [10, 64, 128] \\
& 	 \textit{nb} & [1] \\
\midrule
HAL  & \textit{lr} & [0.03, 0.1, 0.3] \\
& 	 \textit{$\lambda$} & [0.1, 0.2] \\
& 	 \textit{$\beta$} & [0.3, 0.5] \\
& 	 \textit{$\gamma$} & [0.1] \\
\midrule
DER  & \textit{lr} & [0.1, 0.2] \\
& 	 \textit{$\alpha$} & [0.5, 1.0] \\
\midrule
DER++  & \textit{lr} & [0.1, 0.2] \\
& 	 \textit{$\alpha$} & [0.5, 1.0] \\
& 	 \textit{$\beta$} & [0.5, 1.0] \\
\bottomrule
\end{tabular}

\begin{tabular}{ccc}
\toprule
\multicolumn{3}{c}{\textbf{Rotated MNIST}}\\
\textit{Method} & \textit{Par} & \textit{Values} \\
\midrule
SGD   & \textit{lr} & [0.03, 0.1, 0.2] \\
\midrule
oEWC  & \textit{lr} & [0.01, 0.1] \\
& 	  \textit{$\lambda$} & [0.1, 0.7, 1, 10, 30, 90,100] \\
& 	  \textit{$\gamma$} & [0.9, 1.0] \\
\midrule
SI  & \textit{lr} & [0.01, 0.1] \\
& 	  \textit{c} & [0.5, 1.0] \\
& 	  \textit{$\xi$} & [0.9, 1.0] \\
\midrule
ER  & \textit{lr} & [0.1, 0.2] \\
\midrule
GEM  & \textit{lr} & [0.01, 0.3, 0.1] \\
&  \textit{$\gamma$} & [0.1, 0.5, 1]\\
\midrule
A-GEM & \textit{lr} & [0.01, 0.1, 0.3] \\
\midrule
GSS  & \textit{lr} & [0.03, 0.1, 0.2] \\
& 	 \textit{gmbs} & [10, 64, 128] \\
& 	 \textit{nb} & [1] \\
\midrule
HAL  & \textit{lr} & [0.03, 0.1, 0.3] \\
& 	 \textit{$\lambda$} & [0.1, 0.2] \\
& 	 \textit{$\beta$} & [0.3, 0.5] \\
& 	 \textit{$\gamma$} & [0.1] \\
\midrule
DER & \textit{lr} & [0.1, 0.2] \\
& 	 \textit{$\alpha$} & [0.5, 1.0] \\
\midrule
DER++ & \textit{lr} & [0.1, 0.2] \\
& 	  \textit{$\alpha$} & [0.5, 1.0] \\
& 	  \textit{$\beta$} & [0.5, 1.0] \\
\bottomrule
\end{tabular}

\begin{tabular}{cccc}
\toprule
\multicolumn{3}{c}{\textbf{Sequential Tiny ImageNet}}\\
\midrule
\textit{Method} & \textit{Par} &
\textit{Values} \\
\midrule
SGD  & \textit{lr} & [0.01, 0.03, 0.1] \\
\midrule
oEWC & \textit{lr} & [0.01, 0.03] \\
& \textit{$\lambda$} & [10, 25, 30, 90, 100] \\
& \textit{$\gamma$} & [0.9, 0.95, 1.0] \\
\midrule
SI   & \textit{lr} & [0.01, 0.03] \\
& \textit{c} & [0.5] \\
& \textit{$\xi$} & [1.0] \\
\midrule
LwF  & \textit{lr} & [0.01, 0.03] \\
& \textit{$\alpha$} & [0.3, 1, 3] \\
& \textit{T} & [2.0, 4.0] \\
& \textit{wd} & [0.00005, 0.00001] \\
\midrule
PNN  & \textit{lr} & [0.03, 0.1] \\
\midrule
ER  & \textit{lr} & [0.01, 0.03, 0.1] \\
\midrule
A-GEM  & \textit{lr} & [0.003, 0.01] \\
\midrule
iCaRL & \textit{lr} & [0.01, 0.03, 0.1] \\
& \textit{wd} & [0.00005, 0.00001] \\
\midrule
FDR  & \textit{lr} & [0.01, 0.03, 0.1] \\
& \textit{$\alpha$} & [0.03, 0.1, 0.3, 1.0, 3.0] \\
\midrule
DER  & \textit{lr} & [0.01, 0.03, 0.1] \\
& \textit{$\alpha$} & [0.1, 0.5, 1.0] \\
\midrule
DER++ & \textit{lr} & [0.01, 0.03] \\
& \textit{$\alpha$} & [0.1, 0.2, 0.5, 1.0] \\
& \textit{$\beta$} & [0.5, 1.0] \\
\bottomrule
\end{tabular}
    \end{multicols}
\end{table}
\newpage
\begin{table}[H]
    \begin{multicols}{3}
        \setlength{\tabcolsep}{3.5pt}
        \scriptsize
        \begin{tabular}{ccc}
\toprule
\multicolumn{3}{c}{\textbf{Sequential MNIST}}\\
\textit{Method}  & \textit{Par} &
\textit{Values} \\
\midrule
SGD   & \textit{lr} & [0.01, 0.03, 0.1] \\
\midrule
oEWC  & \textit{lr} & [0.01, 0.03, 0.1] \\
& 	  \textit{$\lambda$} & [10, 25, 30, 90, 100] \\
& 	  \textit{$\gamma$} & [0.9, 1.0] \\
\midrule
SI  & \textit{lr} & [0.01, 0.03, 0.1] \\
& 	  \textit{c} & [0.3, 0.5, 0.7, 1.0] \\
& 	  \textit{$\xi$} & [0.9, 1.0] \\
\midrule
LwF   & \textit{lr} & [0.01, 0.03, 0.1] \\
& 	  \textit{$\alpha$} & [0.3, 0.5, 0.7, 1.0] \\
& 	  \textit{T} & [2.0, 4.0] \\
\midrule
PNN   & \textit{lr} & [0.01, 0.03, 0.1] \\
\midrule
ER  & \textit{lr} & [0.03, 0.01, 0.1] \\
\midrule
MER  & \textit{lr} & [0.03, 0.1] \\
& 	  \textit{$\beta,\gamma$} & [1.0] \\
& 	  \textit{nb} & [1] \\
\midrule
GEM & \textit{lr} & [0.01, 0.03, 0.1] \\
&  \textit{$\gamma$} & [0.5, 1]\\
\midrule
A-GEM  & 	  \textit{lr} & [0.03, 0.1] \\
\midrule
iCaRL & \textit{lr} & [0.03, 0.1] \\
& 	  \textit{wd} & [0.0001, 0.0005] \\
\midrule
FDR   & \textit{lr} & [0.03, 0.1] \\
& 	  \textit{$\alpha$} & [0.2, 0.5, 1.0] \\
\midrule
GSS   & \textit{lr} & [0.03, 0.1] \\
& 	  \textit{gmbs} & [10] \\
& 	  \textit{nb} & [1] \\
\midrule
HAL   & \textit{lr} & [0.03, 0.1, 0.2] \\
& 	  \textit{$\lambda$} & [0.1, 0.5] \\
& 	  \textit{$\beta$} & [0.2, 0.5, 0.7] \\
& 	  \textit{$\gamma$} & [0.1, 0.5] \\
\midrule
DER   & \textit{lr} & [0.03, 0.1] \\
& 	  \textit{$\alpha$} & [0.2, 0.5, 1.0] \\
\midrule
DER++ & \textit{lr} & [0.03, 0.1] \\
& 	  \textit{$\alpha$} & [0.2, 0.5, 1.0] \\
& 	  \textit{$\beta$} & [0.2, 0.5, 1.0] \\
\bottomrule
\end{tabular}

\begin{tabular}{cccc}
\toprule
\multicolumn{3}{c}{\textbf{Sequential CIFAR-10}}\\
\textit{Method} & \textit{Par} &
\textit{Values} \\
\midrule
SGD  & \textit{lr} & [0.01, 0.03, 0.1] \\
\midrule
oEWC & \textit{lr} & [0.03, 0.1] \\
& \textit{$\lambda$} & [10, 25, 30, 90, 100]\\
& \textit{$\gamma$} & [0.9, 1.0] \\
\midrule
SI   & \textit{lr} & [0.01, 0.03, 0.1] \\
& \textit{c} & [0.5, 0.1] \\
& \textit{$\xi$} & [1.0] \\
\midrule
LwF  & \textit{lr} & [0.01, 0.03, 0.1] \\
& \textit{$\alpha$} & [0.3, 1, 3, 10] \\
& \textit{T} & [2.0] \\
\midrule
PNN  & \textit{lr} & [0.03, 0.1] \\
\midrule
ER  & \textit{lr} & [0.01, 0.03, 0.1] \\
\midrule
GEM  & \textit{lr} & [0.01, 0.03, 0.1] \\
& \textit{$\gamma$} & [0.5, 1]\\
\midrule
A-GEM & \textit{lr} & [0.03, 0.1] \\
\midrule
iCaRL & \textit{lr} & [0.01, 0.03, 0.1] \\
& \textit{wd} & [0.0001, 0.0005] \\
\midrule
FDR  & \textit{lr} & [0.01, 0.03, 0.1] \\
& \textit{$\alpha$} & [0.03, 0.1, 0.3, 1.0, 3.0] \\
\midrule
GSS  & \textit{lr} & [0.01, 0.03, 0.1] \\
& \textit{gmbs} & [32] \\
& \textit{nb} & [1] \\
\midrule
HAL   & \textit{lr} & [0.01, 0.03, 0.1] \\
& 	  \textit{$\lambda$} & [0.1, 0.5] \\
& 	  \textit{$\beta$} & [0.2, 0.3, 0.5] \\
& 	  \textit{$\gamma$} & [0.1] \\
\midrule
DER  & \textit{lr} & [0.01, 0.03, 0.1] \\
& \textit{$\alpha$} & [0.2, 0.5, 1.0] \\
\midrule
DER++ & \textit{lr} & [0.01, 0.03, 0.1] \\
& \textit{$\alpha$} & [0.1, 0.2, 0.5] \\
& \textit{$\beta$} & [0.5, 1.0] \\
\bottomrule
\end{tabular}

\begin{tabular}{ccc}
\toprule
\multicolumn{3}{c}{\textbf{MNIST-360}}\\
\textit{Method}  & \textit{Par} &
\textit{Values} \\
\midrule
SGD   & \textit{lr} & [0.1, 0.2] \\
& 	  \textit{bs} & [1, 4, 8, 16] \\
\midrule
ER  & \textit{mbs} & [16, 64, 128] \\
& 	  \textit{lr} & [0.1, 0.2] \\
& 	  \textit{bs} & [1, 4, 8, 16] \\
\midrule
MER   & \textit{mbs} & [128] \\
& 	  \textit{lr} & [0.1, 0.2] \\
& 	  \textit{$\beta$} & [1.0] \\
& 	  \textit{$\gamma$} & [1.0] \\
& 	  \textit{nb} & [1] \\
\midrule
A-GEM-R  & \textit{mbs} & [16, 64, 128] \\
& 	  \textit{lr} & [0.1, 0.2] \\
& 	  \textit{bs} & [1, 4, 16] \\
\midrule
GSS   & \textit{mbs} & [16, 64, 128] \\
& 	  \textit{lr} & [0.1, 0.2] \\
& 	  \textit{bs} & [1, 4, 8, 16] \\
& 	  \textit{gmbs} & [16, 64, 128] \\
&     \textit{nb} & [1] \\
\midrule
DER   & \textit{mbs} & [16, 64, 128] \\
& 	  \textit{lr} & [0.1, 0.2] \\
& 	  \textit{bs} & [1, 4, 8, 16] \\
& 	  \textit{$\alpha$} & [0.5, 1.0] \\
\midrule
DER++ & \textit{mbs} & [16, 64, 128] \\
& 	  \textit{lr} & [0.1, 0.2] \\
& 	  \textit{bs} & [1, 4, 8, 16] \\
& 	  \textit{$\alpha$} & [0.2, 0.5] \\
& 	  \textit{$\beta$} & [0.5, 1.0] \\
\bottomrule
\end{tabular}

    \end{multicols}
\caption{Hyperparameter space for Grid-Search}
    \label{tab:hyperparam_space}
\end{table}
}

\bibliographystyle{named}
\bibliography{references}